%% file: main.tex
\documentclass[journal ]{new-aiaa}
\usepackage[utf8]{inputenc}
\usepackage{textcomp}

\usepackage{graphicx}
\usepackage{amsmath}
\usepackage[version=4]{mhchem}
\usepackage{siunitx}
\usepackage{longtable,tabularx}
\usepackage{wasysym}
\usepackage{algorithm2e}
\usepackage{multicol}
\usepackage{multirow}
\RestyleAlgo{ruled}

\title{GNSS-based Lunar Orbit and Clock Estimation With Stochastic Cloning UD Filter}

\author{Keidai Iiyama \footnote{Ph.D. Candidate, Department of Aeronautics and Astronautics, 496 Lomita Mall, Stanford, CA, AIAA Student Member.} and Grace Gao.\footnote{Associate Professor, Department of Aeronautics and Astronautics, 496 Lomita Mall, Stanford, CA.}}
\affil{Stanford University, Stanford, CA, 94305}

\begin{document}

\maketitle

\input{macros}

\input{sections/0_abstract}
\input{sections/1_introduction}
\input{sections/2_0_gnss_navigation}
\input{sections/3_0_udu_filtering}
\input{sections/4_pipeline}

\input{sections/5_simulation_setup}

\input{sections/6_meas_error}
\input{sections/7_result}
\input{sections/8_conclusion}

\section*{Acknowledgments}
This work was supported in part by funding from the the Nakajima Foundation. 
The authors would like to thank Kaila Coimbra, Adam Dai, and members of the Stanford NAV Lab for valuable discussions and technical feedback, and Rebecca Wang and Todd Walter for their advice on GNSS ephemeris error modeling.
AI models (ChatGPT, Gemini) were used to assist with language editing and technical clarity of the manuscript. All derivations, figures, simulations, and conclusions were developed, verified, and validated solely by the authors. 

\bibliography{references}

\end{document}

%% file: macros.tex
\newcommand{\R}{\mathbb{R}}
\newcommand{\SO}{\mathsf{SO}}
\newcommand{\SE}{\mathsf{SE}}

\newcommand{\Phat}{\hat{\mathbf{P}}}
\newcommand{\Pminus}{\hat{\mathbf{P}}^{-}}
\newcommand{\Pplus}{\hat{\mathbf{P}}^{+}}
\newcommand{\Uhat}{\hat{\mathbf{U}}}
\newcommand{\UhatT}{\hat{\mathbf{U}}^{\top}}
\newcommand{\Uhatminus}{\hat{\mathbf{U}}^{-}}
\newcommand{\UhatminusT}{\hat{\mathbf{U}}^{-\top}}
\newcommand{\Uhatplus}{\hat{\mathbf{U}}^{+}}
\newcommand{\UhatplusT}{\hat{\mathbf{U}}^{+\top}}
\newcommand{\Dhat}{\hat{\mathbf{D}}}
\newcommand{\Dhatminus}{\hat{\mathbf{D}}^{-}}
\newcommand{\DhatminusT}{\hat{\mathbf{D}}^{-\top}}
\newcommand{\Dhatplus}{\hat{\mathbf{D}}^{+}}
\newcommand{\DhatplusT}{\hat{\mathbf{D}}^{+\top}}

\newcommand{\CN}{C/N$_0$ }
\newcommand{\CNMath}{\text{C/N}_0}

%% file: sections/0_abstract.tex
\begin{abstract}
This paper presents a terrestrial GNSS–based orbit and clock estimation framework for lunar navigation satellites. To enable high-precision estimation under the low-observability conditions encountered at lunar distances, we develop a stochastic-cloning UD-factorized filter and delayed-state smoother that provide enhanced numerical stability when processing precise time-differenced carrier phase (TDCP) measurements. A comprehensive dynamics and measurement model is formulated, explicitly accounting for relativistic coupling between orbital and clock states, lunar time-scale transformations, and signal propagation delays including ionospheric, plasmaspheric, and Shapiro effects.
The proposed approach is evaluated using high-fidelity Monte-Carlo simulations incorporating realistic multi-constellation GNSS geometry, broadcast ephemeris errors, lunar satellite dynamics, and ionospheric and plasmaspheric delay computed from empirical electron density models. Simulation results demonstrate that combining ionosphere-free pseudorange and TDCP measurements achieves meter-level orbit accuracy and sub-millimeter-per-second velocity accuracy, satisfying the stringent signal-in-space error requirements of future Lunar Augmented Navigation Services (LANS).
\end{abstract}

%% file: sections/1_introduction.tex
\section{Introduction}
\label{sec:introduction}

Lunar exploration has been a topic of significant interest in recent years, driven by international efforts to establish a sustainable presence on the Moon. 
To meet the navigation and timing needs of these missions, the National Aeronautics and Space Administration (NASA), European Space Agency (ESA), and Japan Aerospace Exploration Agency (JAXA) have proposed the Lunar Augmented Navigation Service (LANS) as part of LunaNet~\cite{Israel2020}, the lunar counterpart to Earth's Global Navigation Satellite System (GNSS). 
The LANS aims to provide precise positioning, navigation, and timing (PNT) services to assets on the lunar surface and in orbit by leveraging a constellation of lunar satellites that broadcast Augmented Forward Signals (AFS)~\cite{LNISv5}. 
Specifically, the National Aeronautics and Space Administration (NASA), the European Space Agency (ESA), and the Japan Aerospace Exploration Agency (JAXA) plan to deploy the Lunar Communications Relay and Navigation Systems (LCRNS)~\cite{gramling2024}, the Lunar Communications and Navigation Service (LCNS)~\cite{traveset2024}, and the Lunar Navigation Satellite System (LNSS)~\cite{Murata2022}, respectively.

The central challenge in realizing the LANS is the accurate orbit and clock determination of the lunar navigation satellites themselves. 
The LunaNet Interoperability Specification (LNIS)~\cite{LNISv5} requires lunar navigation satellites to achieve a signal-in-space error (SISE) of less than 40 meters (95th percentile) for position and \SI{10}{mm/s} (95th percentile) for velocity, which includes errors in predicted orbit and clocks, ephemeris fitting, and unmodeled delays in the transmitter.
For LCRNS, the requirements are even more stringent, demanding a position SISE of less than 13.43 meters ($3\sigma$) and a velocity SISE of \SI{1.2}{mm/s} ($3\sigma$)~\cite{gramling2024}.
Unlike terrestrial GNSS satellites, which rely on ground-based monitoring stations for orbit and clock estimation, lunar satellites face significant hurdles due to the absence of such infrastructure on the lunar surface and the limited visibility and availability of Earth-based tracking stations. 

\begin{figure}[ht!]
    \centering
    \includegraphics[width=0.5\textwidth]{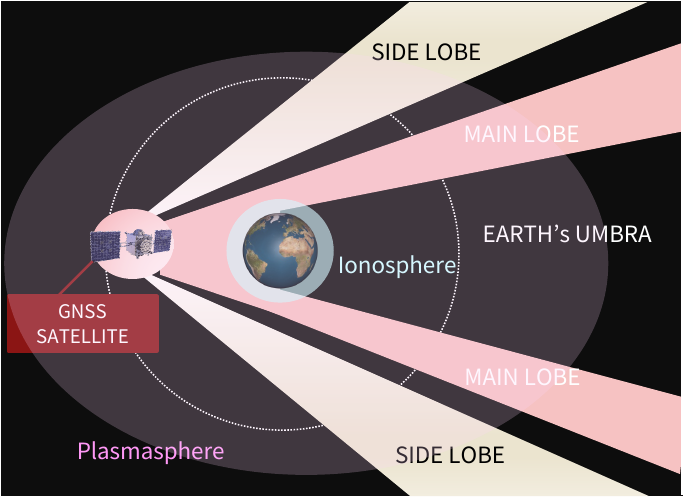}
    \caption{GNSS sidelobe and portion of mainlobe signals spilling into cislunar space (not to scale)}
    \label{fig:gnss_sidelobe}    
\end{figure}

To address these challenges, the use of terrestrial GNSS signals for lunar satellite navigation has been proposed for both LCRNS~\cite{Mina2025LCRNS} and LNSS~\cite{Murata2022}. By tracking the GNSS sidelobes and the portion of mainlobe signals that spill into cislunar space, as shown in Figure~\ref{fig:gnss_sidelobe}, lunar navigation satellites can estimate their orbits and clocks autonomously, reducing reliance on Earth-based tracking.

Initial research on terrestrial GNSS-based lunar orbit and clock estimation proposed both snapshot least-squares and Extended Kalman Filtering (EKF) approaches to estimate satellite states using pseudorange measurements~\cite{Capuano2014EarthMoonGNSS, Capuano2017}, and were verified for feasibility through simulations.
Recently, the Lunar GNSS Receiver Experiment (LuGRE) mission demonstrated this concept in flight~\cite{parker2022lugre, Konitzer2022LuGRE}. The mission successfully tracked terrestrial GNSS signals from lunar orbit and performed onboard estimation using pseudorange measurements.
However, the resulting positioning accuracy was limited to the order of kilometers~\cite{insidegnss2025lugre}. 
This limitation stemmed primarily from the short duration of the experiment (one-hour observation window), and the relatively low stability of the clock compared to navigation satellites (Voltage Controlled Temperature Compensated Crystal Oscillator (VCTCXO)~\cite{Konitzer2022LuGRE}), which prevented the accumulation of sufficient data. 
Due to the poor geometry of GNSS satellites at lunar distances, the filter requires longer durations to converge to a stable solution. 
The reliance on pseudorange measurements further limits accuracy because their noise is higher than that of carrier-phase measurements.

To achieve higher accuracy, \cite{Iiyama2024TDCPNavigation} and \cite{Mina2025LCRNS} proposed utilizing time-differenced carrier phase (TDCP) measurements alongside pseudorange data. 
However, because TDCP measurements depend on both the current and previous states (delayed-state measurements), the standard Kalman filter assumptions are violated. Specifically, the conditional independence assumption
\begin{equation}
    \text{COV} (Z_k | X_k, X_{k-1}) = \text{COV} (Z_k | X_k)
\end{equation}
does not hold for measurements $Z_k$ that depend on both the current state $X_k$ and a past state $X_{k-1}$. 
To handle this, \cite{Iiyama2024TDCPNavigation} employed stochastic cloning, a method that was originally proposed in \cite{Roumeliotis2002StochasticCloning}, while \cite{Mina2025LCRNS} utilized a delayed-state EKF.
These studies demonstrated that incorporating TDCP measurements improves orbit and clock estimation accuracy by 10--30\% compared to pseudorange-only solutions, achieving sub-10 meter accuracy in simulation.

Despite these advancements, existing methods remain limited. First, the Kalman filters used in prior works risk divergence due to the low observability of the system, particularly when processing very precise TDCP measurements.
In such scenarios, the covariance matrix in standard Kalman filters can lose its positive semi-definiteness (PSD) due to numerical errors, leading to filter divergence and degraded performance~\cite{Tapley2004}.
In aerospace applications, it is standard practice to use square-root formulations~\cite{Potter1963StatFiltering} or UD-factorization-based Kalman filters~\cite{Thornton1976TriangularCovariance, BiermanThornton1977_KalmanComparison} to mitigate this.
Square-root filters were notably used in the Apollo program to maximize numerical precision within 8-bit limits~\cite{Fraser2019ApolloMemories}.
Alternatively, the UD-filter factorizes the covariance matrix as $P = UDU^{\top}$, where $U$ is a upper triangular matrix with ones on diagonals and $D$ is a diagonal matrix. 
UD-filter avoids square-root operations, making it computationally more efficient, and has been recently adopted in the Orion Navigation System~\cite{Sud2014OrionEFT1Nav}.
These methods replace the covariance matrix with its factorization, ensuring symmetry and positive semi-definiteness by construction~\cite{DSouzaZanetti2019}.
However, to the best of the authors' knowledge, variants of these covariance-factorization filters capable of processing delayed-state measurements such as TDCP have not yet been presented in the literature. 

Second, while prior works have employed stochastic cloning or delayed-state EKF to handle TDCP measurements, the derivation of smoothing algorithms to refine past state estimates using all available data for post-processing has not been addressed. Recursive equations for fixed-interval smoothing~\cite{Rauch1965RTS} cannot be directly applied in the presence of delayed-state measurements, as the Markov assumptions underlying these equations are violated.

Third, the effects of general and special relativity on GNSS signals and the local clocks of lunar satellites have not been adequately addressed in prior GNSS-based lunar orbit determination and clock synchronization (ODTS) works. 
Relativistic effects induce coupling between orbital states and clock bias states, leading to systematic drift (in orders of $mu$ seconds per day) in estimated clock biases if not properly modeled~\cite{seyffert2025}. 
In addition, while the LuGRE mission synchronized its clock to GPS Time (GPST)~\cite{Konitzer2022LuGRE}, future lunar satellites aiming for high-precision PNT services will need to synchronize clocks to lunar-based reference time frames~\cite{Bourgoin_2025, Turyshev2026RelativisticTimescales, Kopeikin_2024}, making accurate conversions between these time scales essential.

Finally, prior works generally assumed that plasmaspheric delays affecting GNSS signals propagating from Earth to lunar orbit above the ionosphere are negligible~\cite{Iiyama2024TDCPNavigation, Mina2025LCRNS}. 
Recent studies by the authors~\cite{Iiyama2025Iono_iongnss, iiyama2025iono_arxiv}, employing ray-tracing simulations with the Global Core Plasma Model (GCPM)~\cite{Gallagher2000GCPM}, demonstrate that these delays can reach several meters to over ten meters depending on geomagnetic and solar conditions. 
Such errors are non-negligible at the accuracy levels targeted by LANS and must be explicitly modeled and mitigated.

To overcome these limitations, this paper presents a terrestrial GNSS-based lunar ODTS framework that integrates stochastic cloning with a UD-factorized Extended Kalman filter. 
This formulation enhances numerical stability while enabling principled processing of delayed-state TDCP measurements. 
We further derive fixed-interval smoothing equations tailored to delayed-state systems, enabling refined retrospective estimation of orbital and clock states. 
The proposed framework rigorously accounts for relativistic effects on signal propagation and onboard clocks, and employs dual-frequency pseudorange and TDCP measurements to mitigate ionospheric and plasmaspheric delays. 
Performance is evaluated through high-fidelity Monte Carlo simulations incorporating realistic GNSS ephemeris errors, LCRNS orbital dynamics, relativistic corrections, and ionospheric and plasmaspheric delays computed from empirical electron density models of the ionosphere and plasmasphere.

The remainder of this paper is organized as follows. Section \ref{sec:gnss_navigation} describes the dynamics and measurement models used for estimation. Section \ref{sec:udu_filtering} derives the stochastic-cloning UD filter and smoother equations. Section \ref{sec:pipeline} presents the overall orbit and clock estimation pipeline. Section \ref{sec:simulation_setup} details the simulation setup, while Section \ref{sec:meas_error_analysis} discusses the distribution of measurement delays. Section \ref{sec:result} presents the simulation results and performance analysis. Finally, Section \ref{sec:conclusion} concludes the paper and discusses future work.

%% file: sections/2_0_gnss_navigation.tex
\section{Lunar Orbit and Clock Estimation with Terrestrial GNSS}
\label{sec:gnss_navigation}
In this section, we describe the dynamics of the lunar satellite's orbit and clock states, including the effects of time dilation due to relativistic effects. We also discuss how these states are mapped to terrestrial GNSS measurements for orbit and clock estimation, as well as the measurement error sources that must be considered.

\input{sections/2_1_state_dynamics}
\input{sections/2_2_time_clocks}
\input{sections/2_3_gnss_observables}

%% file: sections/2_1_state_dynamics.tex
\subsection{State Definition}
\label{sec:state_definition}

The system state vector to be estimated at the discrete time step $k$ is defined as
\begin{equation}
    \mathbf{x}_k =
    \begin{bmatrix}
        \mathbf{r}_k^{\mathrm{LCRF}} &
        \mathbf{v}_k^{\mathrm{LCRF}} &
        c\,\delta t_k &
        c\,\delta \dot{t}_k &
        c\,\delta \ddot{t}_k &
        \gamma_{\mathrm{srp}}
    \end{bmatrix}^{\!T},
\end{equation}
where $\mathbf{r}_k^{\mathrm{LCRF}}$ and $\mathbf{v}_k^{\mathrm{LCRF}}$ denote the position and velocity of the lunar satellite expressed in the Lunar Celestial Reference Frame (LCRF), which is the realization of the Lunar Celestial Reference System (LCRS). The LCRF is centered at the Moon’s center of mass and shares its orientation with the Barycentric Celestial Reference Frame (BCRF) and Geocentric Celestial Reference Frame (GCRF). 
For notational brevity, the superscript $\mathrm{LCRF}$ is omitted in subsequent sections unless other reference frames are explicitly indicated.

The remaining components of the state vector include the clock bias $\delta t_k$, clock drift $\delta \dot{t}_k$, and clock acceleration $\delta \ddot{t}_k$, all scaled by the speed of light $c$, as well as the solar radiation pressure (SRP) coefficient $\gamma_{\mathrm{srp}}$. 
The clock states are defined with respect to Lunar Coordinate Time (TCL), which serves as the coordinate time of the LCRF. 
A detailed discussion of TCL and its relationship to other time scales is provided in Section~\ref{sec:lunar_time_clock_drifts}.

\subsection{Orbit Dynamics}
\label{sec:orbit_dynamics}
The continuous-time dynamics of the lunar satellite orbit are modeled as
\begin{align}
    \begin{bmatrix} \dot{\mathbf{r}}_k \\ \dot{\mathbf{v}}_k \end{bmatrix}
    &=
    \begin{bmatrix}
        \mathbf{v}_k \\
        \mathbf{a}_g(t_k,\mathbf{r}_k) + \mathbf{a}_{\mathrm{srp}}(t_k,\mathbf{r}_k,\gamma_{\mathrm{srp}})
    \end{bmatrix}
\end{align}
The gravitational acceleration $\mathbf{a}_g(t,\mathbf{r})$ includes contributions from the Moon’s nonspherical gravity field as well as third-body perturbations from the Earth and Sun~\cite{montenbruck2013satellite}:
\begin{align}
    \mathbf{a}_g(t,\mathbf{r})
    &=
    \nabla U_g
    + \sum_{p \in \{\mathrm{Earth},\,\mathrm{Sun}\}}
    \mu_p \left(
    \frac{\mathbf{r}_p - \mathbf{r}}{\|\mathbf{r}_p - \mathbf{r}\|^3}
    -
    \frac{\mathbf{r}_p}{\|\mathbf{r}_p\|^3}
    \right), \\
    U_g
    &=
    \frac{\mu_L}{\|\mathbf{r}\|}
    \sum_{n=0}^{N_{\mathrm{sph}}}
    \sum_{m=0}^{n}
    \left(\frac{R_L}{\|\mathbf{r}\|}\right)^n
    \bar{P}_{nm}(\sin\phi)
    \Big(
        \bar{C}_{nm}\cos m\lambda
        +
        \bar{S}_{nm}\sin m\lambda
    \Big),
    \label{eq:u_g}
\end{align}
where $\mu_L$ is the Moon’s gravitational parameter, $R_L$ is the mean lunar radius, $\lambda$ and $\phi$ are the longitude and latitude in the lunar body-fixed frame, $\bar{P}_{nm}$ are fully normalized associated Legendre polynomials, and $\bar{C}_{nm}$ and $\bar{S}_{nm}$ denote the normalized spherical harmonic coefficients of the lunar gravitational field.

Since the state dynamics are formulated in Lunar Coordinate Time (TCL), all dynamical quantities must be expressed in TCB(TCL)-compatible units. 
The conversion between TDB-compatible and TCB(TCL)-compatible quantities~\cite{Bourgoin_2025, Klioner2008} is given by
\begin{equation}
\begin{aligned}
    \mathcal{X}^{\mathrm{TCL}}
    &=
    \mathcal{X}^{\mathrm{TCB}}
    =
    \frac{1}{1 - L_B}\,
    \mathcal{X}^{\mathrm{TDB}}, \\
    \mu_p^{\mathrm{TCL}}
    &=
    \mu_p^{\mathrm{TCB}}
    =
    \frac{1}{1 - L_B}\,
    \mu_p^{\mathrm{TDB}},
\end{aligned}
\label{eq:tcb_tdb_conversion}
\end{equation}
where $\mathcal{X}$ denotes the scale for positions, and $L_B = 1.550519768 \times 10^{-8}$ is the defining constant relating TCB and TDB. 
This conversion is essential when utilizing planetary ephemerides and gravitational parameters provided in TDB-compatible form.

The acceleration due to solar radiation pressure (SRP), $\mathbf{a}_{\mathrm{srp}}$, is modeled using the cannonball approximation~\cite{Montenbruck2015GNSS}:
\begin{equation}
    \mathbf{a}_{\mathrm{srp}}
    =
    - C_R \frac{A \Phi_S}{c m}
    \left(
    \frac{1\,\mathrm{AU}}{\|\mathbf{r} - \mathbf{r}_S\|}
    \right)^2
    \frac{\mathbf{r}_S - \mathbf{r}}{\|\mathbf{r}_S - \mathbf{r}\|}
    =
    - \gamma_{\mathrm{srp}}
    \left(
    \frac{\Phi_S (1\,\mathrm{AU})^2}{c}
    \right)
    \frac{\mathbf{r}_S - \mathbf{r}}{\|\mathbf{r}_S - \mathbf{r}\|^3},
    \label{eq:srp}
\end{equation}
where $\Phi_S$ denotes the solar flux at 1~AU (1360~\si{W/m^2}), $\mathbf{r}_S$ is the Sun position vector, $C_R = 1 + \epsilon$ is the radiation pressure coefficient with $\epsilon$ denoting the diffuse reflection ratio ($0 \le \epsilon \le 1$), $A$ is the effective cross-sectional area, and $m$ is the spacecraft mass. 
The parameter $\gamma_{\mathrm{srp}} = C_R A / m$ is estimated as part of the state vector.

%% file: sections/2_2_time_clocks.tex
\subsection{Relativistic Effects and Clock Dynamics}
\label{sec:lunar_time_clock_drifts}

Accurate time synchronization among lunar navigation satellites requires a consistent definition of the lunar time reference frame, rigorous modeling of relativistic effects, and a stochastic representation of onboard clock behavior. 
This subsection first introduces the definition of Lunar Coordinate Time (TCL) and Lunar Time (LT), followed by models for relativistic clock drift, gravitational signal delay, and hardware-induced clock noise.

\subsubsection{Lunar Coordinate Time (TCL) and Lunar Time (LT)}
\label{sec:lunar_time_scale}

LANS satellites are expected to broadcast a common lunar time scale for navigation and timing. Although the final definition has not yet been standardized, we assume this reference to be Lunar Time (LT), defined to remain close to the proper time of clocks located on the selenoid, as assumed in \cite{Bourgoin_2025, Turyshev2026RelativisticTimescales}. 
Before introducing LT, we first review the definition of TCL, following~\cite{Kopeikin_2024}.

TCL is the relativistic coordinate time associated with the LCRS, analogous to Geocentric Coordinate Time (TCG) for the Geocentric Celestial Reference System (GCRS) and Barycentric Coordinate Time (TCB) for the Barycentric Celestial Reference System (BCRS). 
The transformation between TCL and TCB is defined by accounting for the Moon’s barycentric motion and the external gravitational potential acting on the Moon:
\begin{equation}
    \frac{d\mathrm{TCL}}{d\mathrm{TCB}} =
    1 - \frac{1}{c^2}
    \left(
        \frac{\|\mathbf{v}_m\|^2}{2}
        +
        U_{\mathrm{ext}}(\mathbf{r}_m)
    \right)
    + \mathcal{O}\!\left(\frac{1}{c^4}\right),
\end{equation}
where $\mathbf{r}_m$ and $\mathbf{v}_m$ denote the Moon’s barycentric position and velocity, respectively, and $U_{\mathrm{ext}}(\mathbf{x}_m)$ represents the gravitational potential of all bodies excluding the Moon, evaluated at the Moon’s location.

For time transfer applications involving Earth-based clocks, it is convenient to express TCL relative to TCG. At first post-Newtonian order, the transformation derived in~\cite{Kopeikin_2024} is given by
\begin{equation}
\begin{aligned}
    \mathrm{TCL}
    &=
    \mathrm{TCG}
    -
    \frac{1}{c^2}
    \int_{t_0}^{t}
    \Bigg\{
        \frac{v_{LE}^2}{2}
        -
        \frac{\mu_E}{r_{LE}}
        -
        \frac{2\mu_E}{r_{LE}} 
        +
        \frac{3}{2}
        \frac{\mu_S}{r_{ES}}
        \left[
            (\mathbf{r}_{ES} \cdot \mathbf{r}_{LE})^2
            -
            \frac{1}{3} r_{ES}^2 r_{LE}^2
        \right]
    \Bigg\}
    dt
    -
    \frac{1}{c^2}
    \left(
        \mathbf{v}_{LE} \cdot \mathbf{r}_{ES}
        -
        \mathbf{v}_{LE} \cdot \mathbf{r}_{LE}
    \right),
    \label{eq:tcl_tcg}
\end{aligned}
\end{equation}
where $\mu_A$ denotes the gravitational parameter of body $A$, and $\mathbf{r}_{AB}$ and $\mathbf{v}_{AB}$ represent the relative position and velocity vectors of body $A$ with respect to body $B$. Subscripts $L$, $E$, and $S$ correspond to the Moon, Earth, and Sun, respectively. Note that the time $t$ is expressed in TCG.

\begin{figure}[ht!]
    \centering
    \includegraphics[width=0.8\textwidth]{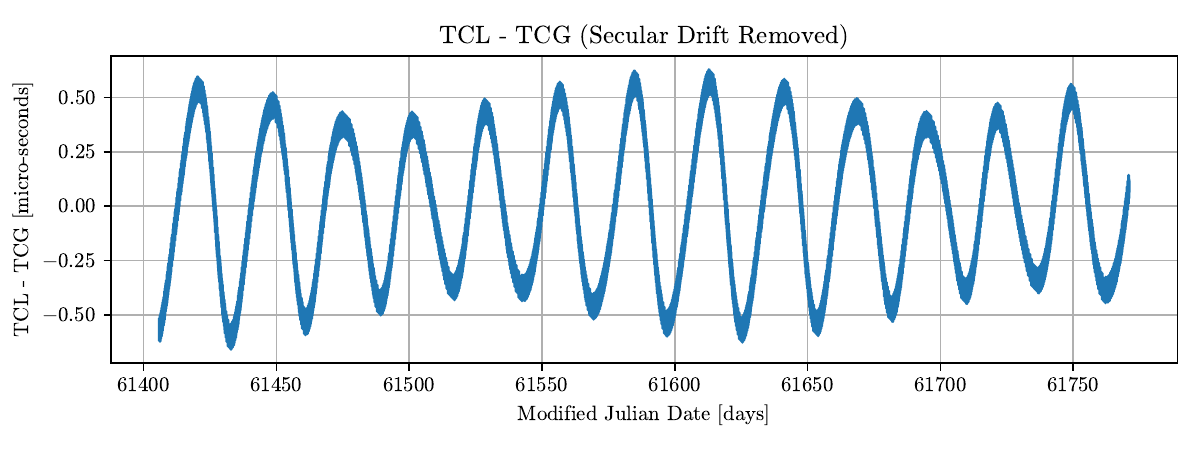}
    \caption{Numerically integrated difference between TCL and TCG for 2027/01/01–2027/12/31. The secular drift component is removed.}
    \label{fig:tcl_tcg}
\end{figure}

Figure~\ref{fig:tcl_tcg} illustrates the numerically integrated difference $\mathrm{TCL}-\mathrm{TCG}$ over the interval 2027/01/01–2027/12/31, using JPL planetary ephemerides DE440 ~\cite{Park2021DE440} and a reference epoch $t_0$ of 1977 January 1, 00:00:00 TAI. The secular drift of $-1.4769~\mu$s/day has been removed to highlight periodic variations.

Using these relations, terrestrial GNSS time (GPST) can be transformed into LT via the sequence~\cite{IERS2010, Kopeikin_2024, Turyshev2026RelativisticTimescales}:
\begin{equation}
\begin{aligned}
    \begin{cases}
    \mathrm{TT} = \mathrm{TAI} + 32.184~\mathrm{s} = (\mathrm{GPST} + 19~\mathrm{s}) + 32.184~\mathrm{s}, \\
    \mathrm{TCG} = \mathrm{TT} + \frac{L_G}{1 - L_G} (\mathrm{TT} - T_0), \\
    \mathrm{TCL} = \mathrm{TCG} + \Delta(\mathrm{TCL}-\mathrm{TCG}), \\
    \mathrm{LT}  = \mathrm{TCL} - L_L (\mathrm{TCL} - T_{L0}),
    \end{cases}
    \label{eq:time_transformation}
\end{aligned}
\end{equation}
where $L_G = 6.969290134 \times 10^{-10}$ and $L_L$ are linear scale constants relating TCG to TT and TCL to LT, respectively, and
$T_0 = 2443144.5003725$ JD (Julian Date) = 1977-01-01 00:00:32.184 TAI is the reference epoch. The term $\Delta(\mathrm{TCL}-\mathrm{TCG})$ is given by~\eqref{eq:tcl_tcg}. 
The coefficient $L_L = \Phi_L/c^2$ is depends on the reference chosen lunar potential $\Phi_L$. 
Reference~\cite{Kopeikin_2024} adopts a lunar gravitational potential of $\Phi_L = 2.822336927 \times 10^{6},\mathrm{m}^2/\mathrm{s}^2$, yielding $L_L = 3.14027 \times 10^{-11}$, whereas \cite{Turyshev2026RelativisticTimescales} assumes $\Phi_L = 2.82123744381 \times 10^{6},\mathrm{m}^2/\mathrm{s}^2$, resulting in $L_L = 3.13905 \times 10^{-11}$. The definitive value of $L_L$ has not yet been established through international standardization, and the initial lunar timescale offset, $T_{L0}$, likewise remains undefined.

Efficient onboard implementation of the TCL--TCG transformation requires compact analytical approximations of $\Delta(\mathrm{TCG}-\mathrm{TCL})$. While global polynomial approximations are challenging to maintain over long durations~\cite{Lu2025_LTE440}, piecewise polynomial or spline-based representations over short intervals may offer a practical compromise between accuracy and computational efficiency. 
It would also be necessary to represent TCL--TCG in TCL (Eq \eqref{eq:tcl_tcg} is in TCG), to execute inverse transformation without root-solving.

\subsubsection{Relativistic Clock Drift with respect to TCL}

Beyond coordinate time transformations, the proper time of the onboard clock experiences relativistic drift relative to TCL due to both gravitational and kinematic time dilation. The resulting rate difference is given by~\cite{seyffert2025}
\begin{equation}
    \dot{\delta} t_{\mathrm{rel}} (\mathbf{r}, \mathbf{v})
    =
    \frac{d(\tau - \mathrm{TCL})}{d\mathrm{TCL}}
    =
    \frac{1}{c^2}
    \left(
        U_L(\mathbf{r})
        +
        \frac{v^2}{2}
    \right)
    \approx
    \frac{1}{c^2}
    \left(
        \frac{\mu_L}{r}
        +
        \frac{v^2}{2}
    \right),
    \label{eq:relativistic_delay_orbit}
\end{equation}
where $\tau$ denotes the satellite proper time, $U_L(\mathbf{r})$ is the lunar gravitational potential at the spacecraft location $\mathbf{r}$, $r$ is the lunicentric distance, and $\mathbf{v}, v$ is the orbital velocity and its norm in the LCRS.

\subsubsection{Shapiro Time Delay}

Signal propagation through curved spacetime introduces an additional relativistic delay known as the Shapiro effect. Considering solar gravity as the dominant contributor, the time delay for a GNSS signal transmitted from satellite $i$ to the lunar receiver is~\cite{Shapiro1964TimeDelay}
\begin{equation}
    \Delta t^s_{k,i}
    =
    \frac{2\mu_S}{c^3}
    \ln
    \left(
    \frac{
        r_k^{\mathrm{BCRF}} + r_{tx,i}^{\mathrm{BCRF}} + R_{k,i}
    }{
        r_k^{\mathrm{BCRF}} + r_{tx,i}^{\mathrm{BCRF}} - R_{k,i}
    }
    \right),
    \label{eq:shapiro_delay}
\end{equation}
where $\mu_S$ is the solar gravitational parameter, $r_k^{\mathrm{BCRF}}$ and $r_{tx,i}^{\mathrm{BCRF}}$ are the barycentric distances of the receiver and transmitter from the Sun, respectively, and $R_{k,i}$ is the Euclidean distance between them. For Earth-Moon links, the Shapiro delays from the Sun is in \SI{20}{ns} to \SI{30}{ns}, which dominates the delay when compared with \SI{0.1}{ns} to \SI{0.2}{ns} from Earth, and \SI{1}{ps} to \SI{3}{ps} from the Moon~\cite{Turyshev2026RelativisticTimescales}.

\subsubsection{Stochastic Clock Drift Model}

In addition to relativistic effects, onboard clock behavior is influenced by oscillator imperfections. This effect is modeled as a stochastic process. The discrete-time clock dynamics are expressed as~\cite{Zucca2005}
\begin{equation}
\begin{aligned}
    \begin{bmatrix}
        c\delta t_k \\
        c\delta \dot{t}_k \\
        c\delta \ddot{t}_k
    \end{bmatrix}
    &=
    \begin{bmatrix}
        1 & \Delta t & \frac{1}{2}\Delta t^2 \\
        0 & 1 & \Delta t \\
        0 & 0 & 1
    \end{bmatrix}
    \begin{bmatrix}
        c\delta t_{k-1} \\
        c\delta \dot{t}_{k-1} \\
        c\delta \ddot{t}_{k-1}
    \end{bmatrix}
    +
    \begin{bmatrix}
        c \int_{t_{k-1}}^{t_k} \dot{\delta} t_{\mathrm{rel}} (\mathbf{r}(t'), \mathbf{v}(t')) dt' \\
        c \dot{\delta} t_{\mathrm{rel}} (\mathbf{r}(t_k), \mathbf{v}(t_k))\\
        0
    \end{bmatrix}
    +
    \boldsymbol{\epsilon}_k^{\mathrm{clk}},
\end{aligned}
\end{equation}
where $\Delta t = t_k - t_{k-1}$ and $\boldsymbol{\epsilon}_k^{\mathrm{clk}} \sim \mathcal{N}(0, Q_k^{\mathrm{clk}})$ represents the clock process noise. This formulation explicitly couples clock and orbital dynamics through the relativistic drift term $\dot{\delta} t_{\mathrm{rel}} ((\mathbf{r}(t_k), \mathbf{v}(t_k))$.

The process noise covariance is modeled as~\cite{Zucca2005}
\begin{equation}
    Q^{\mathrm{clk}}_k =
    c^2
    \begin{bmatrix}
        q_1 \Delta t + \frac{q_2 \Delta t^3}{3} + \frac{q_3 \Delta t^5}{20}
        &
        \frac{q_2 \Delta t^2}{2} + \frac{q_3 \Delta t^4}{8}
        &
        \frac{q_3 \Delta t^3}{6}
        \\
        \frac{q_2 \Delta t^2}{2} + \frac{q_3 \Delta t^4}{8}
        &
        q_2 \Delta t + \frac{q_3 \Delta t^3}{3}
        &
        \frac{q_3 \Delta t^2}{2}
        \\
        \frac{q_3 \Delta t^3}{6}
        &
        \frac{q_3 \Delta t^2}{2}
        &
        q_3 \Delta t
    \end{bmatrix},
    \label{eq:clock_process_noise}
\end{equation}
where $q_1$, $q_2$, and $q_3$ are the diffusion coefficients corresponding to phase noise, frequency noise, and frequency drift (aging), respectively.

%% file: sections/2_3_gnss_observables.tex
\subsection{GNSS Observables}
\label{sec:gnss_observables}

\subsubsection{Pseudorange and Carrier-Phase Measurements}
\label{sec:pseudorange_carrierphase}

Consider a lunar receiver tracking signal $L$ from GNSS satellite $i$ (constellation $S$) at discrete epoch $t_k$. 
The code pseudorange and carrier-phase measurements can be written as~\cite{kaplan2017understanding}
\begin{align}
    \rho_{i,L}(t_k)
    &=
    \bar{\rho}_{i}(t_k)
    + I_{i,L}^{P}(t_k)
    + c\,\Delta t^{S}_{k,i}
    + \delta P_{i,L}(t_k)
    + w_{\rho,k},
    \label{eq:pseudorange}
    \\
    \Phi_{i,L}(t_k)
    &=
    \bar{\rho}_{i}(t_k)
    + I_{i,L}^{C}(t_k)
    + c\,\Delta t^{S}_{k,i}
    + \lambda_L N_{i,L}
    + \delta C_{i,L}(t_k)
    + w_{\Phi,k},
    \label{eq:carrier_phase}
\end{align}
where $\bar{\rho}_i(t_k)$ is the nominal geometric-plus-clock range (defined below), $I_{i,L}^{P}$ and $I_{i,L}^{C}$ denote ionospheric/plasmaspheric contributions to code and carrier phase, respectively, $\Delta t^S_{k,i}$ is the Shapiro delay expressed in time, $N_{i, L}$ is the integer ambiguity, $\delta P_{i,L}$ and $\delta C_{i,L}$ collect instrumental and modeling biases, and $w_{\rho,k}$ and $w_{\Phi,k}$ represent measurement noise.

The nominal range and aggregated bias terms are modeled as
\begin{equation}
\begin{aligned}
    \bar{\rho}_{i}(t_k)
    &=
    \left\|
        \mathbf{r}_k^{\mathrm{ITRF}}
        -
        \bar{\mathbf{r}}_{tx,i}^{\mathrm{ITRF}}(\bar{t}_{tx,i},\boldsymbol{\alpha})
    \right\|
    +
    c\left[
        \delta t_k^{\mathrm{GPST}}
        -
        \bar{\delta t}_{tx,i}^{\mathrm{GPST}}(\bar{t}_{tx,i},\boldsymbol{\alpha})
    \right], \\
    \delta P_{i,L}
    &=
    \left(K^{P,r}_{S,L} - K^{P,t}_{S,i,L}\right)
    + \Delta r_{tx,i}
    + \Delta \mathrm{PCO}_{i,L}
    + M_P, \\
    \delta C_{i,L}
    &=
    \left(K^{C,r}_{S,L} - K^{C,t}_{S,i,L}\right)
    + \Delta r_{tx,i}
    + \Delta \mathrm{PCO}_{i,L}
    + \lambda_L \omega_C
    + M_C,
\end{aligned}
\end{equation}
where $\mathbf{r}_k^{\mathrm{ITRF}}$ is the receiver position in International Terrestrial Reference Frame (ITRF) at $t_k$, $\bar{\mathbf{r}}_{tx,i}^{\mathrm{ITRF}}(\cdot)$ and $\bar{\delta t}_{tx,i}^{\mathrm{GPST}}(\cdot)$ are the broadcast-ephemeris transmitter position and clock bias evaluated at the estimated transmission time $\bar{t}_{tx,i}$ (Sec.~\ref{sec:light_time}), and $\boldsymbol{\alpha}$ denotes the broadcast ephemeris parameters. 
All symbols are summarized in Table~\ref{tab:gnss_symbols}.

\begin{table}[ht!]
    \centering
    \caption{Definitions of Symbols Used in GNSS Measurement Equations}
    \begin{tabular}{ll}
    \hline \hline
    Symbol & Definition \\
    \hline
    $\rho_{i, L}$ & Pseudorange measurement from satellite $i$ using signal $L$ \\
    $\boldsymbol{\alpha}$ & Ephemeris parameters of GNSS satellite \\
    $\bar{t}_{tx, i}$ & Estimated transmission time from satellite $i$ \\
    $\mathbf{r}_k^{ITRF}$ & Receiver position in International Terrestrial Reference Frame (ITRF) at time step $k$ \\
    $\bar{\mathbf{r}}_{tx, i}^{ITRF} (\bar{t}_{tx, i}, \boldsymbol{\alpha})$ & 
    Estimated transmitter satellite position in ITRF at transmission time \\
    $\delta t_k^{GPST}$ & Receiver clock bias in GPS Time (GPST) \\
    $\bar{\delta} t_{tx, i}^{GPST}(\bar{t}_{tx, i}, \boldsymbol{\alpha})$ & Estimated transmitter satellite clock bias in GPST at transmission time \\
    $I_{i, L}^{P}, I_{i, L}^C$ & Ionospheric and plasmaspheric delay on pseudorange (P) and carrier phase (C) measurements \\
    $c \Delta t^S_{i, L}$ & Shapiro time delay in meters \\
    $K^{P, r}_{S, L}, K^{C, r}_{S, L}$ & Receiver code (P) and carrier phase (C) instrumental delay for constellation $S$ and signal $L$ \\
    $K^{P, t}_{S, i, L}, K^{C, t}_{S, i, L}$ & Transmitter code (P) and carrier phase (C) instrumental delay for constellation $S$, satellite $i$, signal $L$ \\   
    $M^P, M^C$ & Multipath error for pseudorange (P) and carrier phase (C) \\
    $N_{i, L}$ & Integer ambiguity for satellite $i$ and signal $L$ \\
    $\Delta r_{tx, i}$ & Ephemeris errors in line-of-sight direction \\
    $\Delta PCO_{i, L}$ & Antenna phase center offset corrections with respect to the reference point in ephemeris \\
    $\lambda_{L}$ & Wavelength of signal $L$ \\
    $\omega_C$ & Carrier phase wind-up due to circular polarization \\
    $w_{\rho, k}$ & Measurement noise at time step $k$ (assumed Gaussian $\sim \mathcal{N}(0, \sigma^2_{\rho, k})$) \\
    \hline \hline
    \end{tabular}
    \label{tab:gnss_symbols}
\end{table}

\subsubsection{Multi-Constellation Inter-System Biases (ISB)}
\label{sec:isb}

The above measurement equations are expressed in ITRF coordinates and GPS Time (GPST). 
For Galileo, the broadcast clock correction is provided in Galileo System Time (GST) using the Galileo ephemeris parameters $\boldsymbol{\alpha}_{\mathrm{GAL}}$. 
We convert the Galileo transmitter clock bias to GPST via
\begin{equation}
    \bar{\delta t}_{tx,i}^{\mathrm{GPST}}
    =
    \bar{\delta t}_{tx,i}^{\mathrm{GST}}(\bar{t}_{tx,i},\boldsymbol{\alpha}_{\mathrm{GAL}})
    - \mathrm{GGTO},
\end{equation}
where $\mathrm{GGTO}$ is the GST-to-GPST offset provided in the Galileo navigation message.

In precise orbit determination with multi-constellation measurements, the system time offset and residual hardware delays are often absorbed into a single inter-system bias (ISB) parameter~\cite{Montenbruck2022Sentinel6RTN}:
\begin{equation}
    B_{\mathrm{GAL}}
    =
    c\,\mathrm{GGTO}
    +
    \left(K^{P,r}_{S,L} - K^{P,t}_{S,i,L}\right).
\end{equation}
In this work, we assume that residual inter-system hardware delays are negligible after calibration compared with other dominant error sources. Estimating $B_{\mathrm{GAL}}$ as a constant or slowly time-varying parameter is left to future work.

\subsubsection{Light-Time Equation}
\label{sec:light_time}
The estimated transmission time $\bar{t}_{tx,i}^{\mathrm{GPST}}$ is obtained by iteratively solving the light-time equation
\begin{equation}
    \bar{t}_{tx,i}^{\mathrm{GPST}}
    =
    t_k^{\mathrm{GPST}}
    -
    \frac{
        \left\|
            \mathbf{r}_k^{\mathrm{ITRF}}
            -
            \bar{\mathbf{r}}_{tx,i}^{\mathrm{ITRF}}(\bar{t}_{tx,i}^{\mathrm{GPST}},\boldsymbol{\alpha})
        \right\|
    }{c},
    \label{eq:light_time_equation}
\end{equation}
where $t_k^{\mathrm{GPST}}$ and $\mathbf{r}_k^{\mathrm{ITRF}}$ are obtained from the lunar time/frame quantities via the transformations in~\eqref{eq:time_transformation}. 
The computation of $\bar{\mathbf{r}}_{tx,i}^{\mathrm{ITRF}}$ from broadcast ephemerides follows~\cite{IS-GPS-200N}.

\subsubsection{Ionospheric and Plasmaspheric Delays and Mitigation}
\label{sec:iono_plasma_mitigation}
GNSS signals propagating from Earth to lunar orbit experience dispersive delays through the ionosphere and plasmasphere. 
The dominant first-order term can be removed by forming an ionosphere-free (IF) combination of dual-frequency measurements. For example, using L1 and L5 signals, the IF pseudorange is~\cite{kaplan2017understanding}
\begin{align}
    \rho_{i,\mathrm{IF}}(t_k)
    &=
    \alpha_{L1}\,\rho_{i,L1}(t_k)
    -
    \alpha_{L5}\,\rho_{i,L5}(t_k)
    \nonumber \\
    &=
    \bar{\rho}_{i}(t_k)
    + \tilde{I}_{i,\mathrm{IF}}^{P}(t_k)
    + c\,\Delta t^{S}_{k,i}
    + \delta P_{i,\mathrm{IF}}(t_k)
    + w_{\rho,\mathrm{IF},k},
    \label{eq:if_pseudorange}
    \\
    w_{\rho,\mathrm{IF},k}
    &\sim
    \mathcal{N}\!\left(0,\,
    \alpha_{L1}^2\sigma_{L1}^2
    +
    \alpha_{L5}^2\sigma_{L5}^2
    \right),
    \label{eq:inflated_noise}
    \\
    \alpha_{L1}
    &=
    \frac{f_1^2}{f_1^2 - f_2^2},
    \qquad
    \alpha_{L5}
    =
    \frac{f_2^2}{f_1^2 - f_2^2},
\end{align}
where $\tilde{I}_{i,\mathrm{IF}}^{P}$ denotes the residual delay after first-order cancellation (e.g., higher-order terms and bending-related contributions), and
\begin{equation}
    \delta P_{i,\mathrm{IF}}
    =
    \alpha_{L1}\,\delta P_{i,L1}
    -
    \alpha_{L5}\,\delta P_{i,L5}.
\end{equation}
As indicated by~\eqref{eq:inflated_noise}, the IF combination increases the effective measurement noise.

\subsubsection{Time-Differenced Carrier Phase Measurements}
\label{sec:tdcp}

Carrier-phase measurements are substantially more precise than pseudorange; however, in the lunar scenario, their direct use is complicated by the coupling between receiver clock bias, propagation delays, and the unknown integer ambiguity. 
To eliminate the integer ambiguity, we form time-differenced carrier phase (TDCP) measurements:
\begin{equation}
    \Delta \Phi_{k,l,i,L}
    =
    \Phi_{i,L}(t_k)
    -
    \Phi_{i,L}(t_l),
    \qquad l < k,
    \label{eq:tdcp}
\end{equation}
which removes the constant ambiguity term $\lambda_L N_{i,L}$ and suppresses slowly varying bias components in $\delta C_{i,L}$. 
TDCP therefore provides a high-precision measurement of relative range change over the interval $[t_l,t_k]$, and serves as a key observable for improving lunar orbit and clock estimation accuracy. 
For this paper, we assume that $l = k-1$, i.e., consecutive epochs, though the proposed filtering and smoothing framework in Section~\ref{sec:udu_filtering} can accommodate arbitrary time differences.

%% file: sections/3_0_udu_filtering.tex
\section{Stochastic-Cloning UD Filtering and Smoothing}
\label{sec:udu_filtering}

This section presents a numerically robust filtering and smoothing framework for processing TDCP measurements in low-observability lunar navigation scenarios. 
While stochastic-cloning EKF have previously been proposed to accommodate delayed-state measurements~\cite{Iiyama2024TDCPNavigation}, we observe that their direct covariance-propagation formulations are susceptible to numerical instability, particularly when processing low-noise TDCP observables. 

To address these limitations, we develop a stochastic-cloning filtering and smoothing algorithm based on UD factorization. By propagating a triangular factorization of the covariance matrix rather than the covariance itself, the proposed formulation preserves symmetry and positive semi-definiteness by construction, significantly improving numerical stability. 
Moreover, we derive fixed-interval smoothing equations compatible with stochastic cloning, enabling retrospective refinement of orbital and clock estimates using all available measurements.

\input{sections/3_1_udu_factorization.tex}
\input{sections/3_2_time_update.tex}
\input{sections/3_3_measurement_update.tex}
\input{sections/3_4_reducing_computation.tex}
\input{sections/3_5_smoothing.tex}

%% file: sections/3_1_udu_factorization.tex
\subsection{UD Factorization of the Covariance Matrix}
\label{sec:udu_factorization}

Let $\hat{\mathbf{x}}_{k|m} \in \mathbb{R}^n$ and $\hat{\mathbf{P}}_{k|m} \in \mathbb{R}^{n \times n}$ denote the state estimate and its associated covariance matrix at discrete time step $k$, conditioned on measurements collected up to time $m$. 
We adopt the UD factorization of the covariance matrix~\cite{BiermanThornton1977_KalmanComparison, Maybeck1979_SMEC}
\begin{equation}
    \hat{\mathbf{P}}_{k|m}
    =
    \mathbf{U}_{k|m}\,\mathbf{D}_{k|m}\,\mathbf{U}_{k|m}^{\top},
\end{equation}
where $\mathbf{U}_{k|m}$ is a unit upper-triangular matrix and $\mathbf{D}_{k|m}$ is a diagonal matrix with strictly positive diagonal entries. 
Instead of directly propagating and updating $\hat{\mathbf{P}}_{k|m}$, the UD filter operates on the factors $\mathbf{U}_{k|m}$ and $\mathbf{D}_{k|m}$, which guarantees symmetry and positive semi-definiteness of the reconstructed covariance by construction, while significantly improving numerical stability and precision.

At the initial time step, the UD factorization of the prior covariance $\hat{\mathbf{P}}_{0|0}$ can be computed using an algorithm closely related to Cholesky factorization, as summarized in Algorithm~\ref{alg:udu_factorization}~\cite{Maybeck1979_SMEC}.

\begin{algorithm}[ht!]
\SetAlgoLined
\LinesNumbered
\KwIn{Symmetric positive-definite covariance matrix $\hat{\mathbf{P}} = [P_{ij}] \in \mathbb{R}^{n \times n}$}
\KwOut{Upper-triangular matrix $\mathbf{U}$ and diagonal matrix $\mathbf{D}$ such that $\hat{\mathbf{P}} = \mathbf{U}\mathbf{D}\mathbf{U}^{\top}$}

\For{$i = 1$ \KwTo $n$}{
    $D_{ii} \gets P_{ii}$\;
}

\For{$i = 1$ \KwTo $n-1$}{
    $U_{i n} \gets P_{i n} / D_{nn}$\;
}
$U_{nn} \gets 1$\;

\For{$j = n$ \KwTo $1$}{
    $D_{jj} \gets P_{jj} - \sum_{k=j+1}^{n} U_{jk}^2 D_{kk}$\;
    $U_{jj} \gets 1$\;
    \For{$i = j-1$ \KwTo $1$}{
        $U_{ij} \gets \frac{1}{D_{jj}}
        \left(
            P_{ij}
            -
            \sum_{k=j+1}^{n} U_{ik} D_{kk} U_{jk}
        \right)$\;
    }
}
\caption{UD factorization of the initial covariance matrix~\cite{Maybeck1979_SMEC}.}
\label{alg:udu_factorization}
\end{algorithm}

%% file: sections/3_2_time_update.tex
\subsection{Augmented-State Time Update}
\label{sec:time_update}

To process TDCP measurements, which depend on both current and past states, we augment the system state by introducing cloned copies of previous states. This approach, known as stochastic cloning~\cite{Roumeliotis2002StochasticCloning}, enables the incorporation of delayed-state measurements within a Kalman filtering framework while preserving statistical consistency.

At time step $k-1$, the augmented state estimate and covariance matrix are initialized as
\begin{equation}
\hat{\mathbf{x}}_{k-1|k-1}^{\mathrm{sc}}
=
\begin{bmatrix}
    \hat{\mathbf{x}}_{k-1|k-1} \\
    \hat{\mathbf{x}}_{k-1|k-1}
\end{bmatrix},
\qquad
\hat{\mathbf{P}}_{k-1|k-1}^{\mathrm{sc}}
=
\begin{bmatrix}
    \hat{\mathbf{P}}_{k-1|k-1} & \hat{\mathbf{P}}_{k-1|k-1} \\
    \hat{\mathbf{P}}_{k-1|k-1} & \hat{\mathbf{P}}_{k-1|k-1}
\end{bmatrix},
\end{equation}
where the first block corresponds to the propagated state and the second block represents the cloned (static) state.

The time-propagated augmented state and covariance at epoch $k$, prior to measurement update, are then given by~\cite{Roumeliotis2002StochasticCloning}
\begin{equation}
\hat{\mathbf{x}}_{k|k-1}^{\mathrm{sc}}
=
\begin{bmatrix}
    \hat{\mathbf{x}}_{k|k-1} \\
    \hat{\mathbf{x}}_{k-1|k-1}
\end{bmatrix},
\end{equation}
\begin{equation}
\hat{\mathbf{P}}_{k|k-1}^{\mathrm{sc}}
=
\begin{bmatrix}
    \mathbf{\Phi}_{k,k-1}\,\hat{\mathbf{P}}_{k-1|k-1}\,\mathbf{\Phi}_{k,k-1}^{\top}
    +
    \mathbf{G}_k\,\mathbf{Q}_k^{D}\,\mathbf{G}_k^{\top}
    &
    \mathbf{\Phi}_{k,k-1}\,\hat{\mathbf{P}}_{k-1|k-1}
    \\
    \hat{\mathbf{P}}_{k-1|k-1}\,\mathbf{\Phi}_{k,k-1}^{\top}
    &
    \hat{\mathbf{P}}_{k-1|k-1}
\end{bmatrix},
\end{equation}
where $\mathbf{\Phi}_{k,k-1}$ is the state transition matrix of the original (non-augmented) system, $\mathbf{Q}_k$ is the process noise covariance, and $\mathbf{G}_k \in \mathbb{R}^{n \times n}$ and $\mathbf{Q}_k^{D} \in \mathbb{R}^{n \times n}$ are obtained via the UD factorization $\mathbf{Q}_k = \mathbf{G}_k \mathbf{Q}_k^{D} \mathbf{G}_k^{\top}$ using Algorithm~\ref{alg:udu_factorization}. 
The state transition matrix $\mathbf{\Phi}_{k,k-1}$ is computed by integrating the variational equations associated with the nonlinear system dynamics.

In the proposed UD filter, instead of propagating the augmented covariance matrix $\hat{\mathbf{P}}_{k|k-1}^{\mathrm{sc}}$ directly, we directly construct the UD factors of $\hat{\mathbf{P}}_{k|k-1}^{\mathrm{sc}}$:
\begin{equation}
    \hat{\mathbf{P}}_{k|k-1}^{\mathrm{sc}}
    =
    \hat{\mathbf{U}}_{k|k-1}^{\mathrm{sc}}
    \,
    \mathbf{D}_{k|k-1}^{\mathrm{sc}}
    \,
    \bigl(\hat{\mathbf{U}}_{k|k-1}^{\mathrm{sc}}\bigr)^{\top},
\end{equation}
where the factors are given as
\begin{align}
    \hat{\mathbf{U}}_{k|k-1}^{\mathrm{sc}}
    &=
    \begin{bmatrix}
        \mathbf{G}_k
        &
        \mathbf{\Phi}_{k,k-1}\,\hat{\mathbf{U}}_{k-1|k-1}
        \\
        \mathbf{0}
        &
        \hat{\mathbf{U}}_{k-1|k-1}
    \end{bmatrix},
    \label{eq:U_prior}
    \\
    \mathbf{D}_{k|k-1}^{\mathrm{sc}}
    &=
    \begin{bmatrix}
        \mathbf{Q}_k^{D} & \mathbf{0} \\
        \mathbf{0} & \hat{\mathbf{D}}_{k-1|k-1}
    \end{bmatrix}.
    \label{eq:D_prior}
\end{align}
using $\hat{\mathbf{U}}_{k-1|k-1}$ and $\hat{\mathbf{D}}_{k-1|k-1}$, the UD factors of the posterior covariance at time step $k-1$.

%% file: sections/3_3_measurement_update.tex
\subsection{Measurement Update}
\label{sec:meas_update}

In the UD filtering framework, measurements at epoch $k$ are processed sequentially. 
Let $\hat{\mathbf{P}}_{k|k_m}^{\mathrm{sc}} \in \mathbb{R}^{2n\times 2n}$ denote the augmented covariance after assimilating the first $m$ measurements at time step $k$ (with $m=0,\ldots,M_k$), where $M_k$ is the number of measurements at epoch $k$. 
By definition, we have
$\hat{\mathbf{P}}_{k|k_0}^{\mathrm{sc}} = \hat{\mathbf{P}}_{k|k-1}^{\mathrm{sc}}$ and $\hat{\mathbf{P}}_{k|k_{M_k}}^{\mathrm{sc}} = \hat{\mathbf{P}}_{k|k}^{\mathrm{sc}}$.

\subsubsection{Sequential covariance update}

For the $m$-th scalar measurement at epoch $k$, with measurement function $h_{k,m}(\cdot)$, Jacobian $\mathbf{H}_{k,m}$, and noise variance $R_{k,m}$, the covariance update in the (augmented) Kalman filter form is
\begin{equation}
\begin{aligned}
    \hat{\mathbf{P}}_{k|k_m}^{\mathrm{sc}}
    &=
    \hat{\mathbf{P}}_{k|k_{m-1}}^{\mathrm{sc}}
    -
    \hat{\mathbf{P}}_{k|k_{m-1}}^{\mathrm{sc}}\mathbf{H}_{k,m}^{\top}
    \left(
        \mathbf{H}_{k,m}\hat{\mathbf{P}}_{k|k_{m-1}}^{\mathrm{sc}}\mathbf{H}_{k,m}^{\top}
        +
        R_{k,m}
    \right)^{-1}
    \mathbf{H}_{k,m}\hat{\mathbf{P}}_{k|k_{m-1}}^{\mathrm{sc}}.
\end{aligned}
\end{equation}

The measurement Jacobian with respect to the stochastic-cloning augmented state $\mathbf{X}^{\mathrm{sc}}=[\mathbf{X}_k^{\top},\mathbf{X}_{k-1}^{\top}]^{\top}$ is
\begin{equation}
\begin{aligned}
    \mathbf{H}_{k,m}
    &=
    \left.\frac{\partial h_{k,m}(\mathbf{X}^{\mathrm{sc}})}{\partial \mathbf{X}^{\mathrm{sc}}}\right|_{\mathbf{X}^{\mathrm{sc}}=\hat{\mathbf{X}}_{k|k_{m-1}}^{\mathrm{sc}}}
    =
    \begin{bmatrix}
        \dfrac{\partial h_{k,m}}{\partial \mathbf{X}_k} &
        \dfrac{\partial h_{k,m}}{\partial \mathbf{X}_{k-1}}
    \end{bmatrix}
    \triangleq
    \begin{bmatrix}
        \mathbf{H}^{(1)}_{k,m} & \mathbf{H}^{(2)}_{k,m}
    \end{bmatrix}
    \in \mathbb{R}^{1\times 2n}.
\end{aligned}
\end{equation}
For measurements that depend only on the current state (e.g., pseudorange), we have $\mathbf{H}^{(2)}_{k,m}=\mathbf{0}$.

\subsubsection{UD rank-one update}

Let the prior augmented covariance before and after processing the $m$-th measurement admit the UD factorization
\begin{equation}
    \hat{\mathbf{P}}_{k|k_{m-1}}^{\mathrm{sc}}
    =
    \hat{\mathbf{U}}^{-}\hat{\mathbf{D}}^{-}\bigl(\hat{\mathbf{U}}^{-}\bigr)^{\top},
    \qquad
    \hat{\mathbf{P}}_{k|k_{m}}^{\mathrm{sc}}
    =
    \hat{\mathbf{U}}^{+}\hat{\mathbf{D}}^{+}\bigl(\hat{\mathbf{U}}^{+}\bigr)^{\top},
\end{equation}
where $(\hat{\mathbf{U}}^{-},\hat{\mathbf{D}}^{-})$ and $(\hat{\mathbf{U}}^{+},\hat{\mathbf{D}}^{+})$ are the UD factors before and after the update, respectively. 
Following~\cite{Thornton1976TriangularCovariance, Maybeck1979_SMEC}, define the innovation variance
\begin{equation}
    a
    =
    \mathbf{H}_{k,m}\hat{\mathbf{P}}_{k|k_{m-1}}^{\mathrm{sc}}\mathbf{H}_{k,m}^{\top}
    +
    R_{k,m},
\end{equation}
and the auxiliary vectors
\begin{equation}
    \mathbf{f} = (\hat{\mathbf{U}}^{-})^{\top}\mathbf{H}_{k,m}^{\top},
    \qquad
    \mathbf{g} = \hat{\mathbf{D}}^{-}\mathbf{f}.
\end{equation}
Then the covariance update can be written as a rank-one downdate on $\hat{\mathbf{D}}^{-}$:
\begin{equation}
\begin{aligned}
    \hat{\mathbf{U}}^{+}\hat{\mathbf{D}}^{+}\bigl(\hat{\mathbf{U}}^{+}\bigr)^{\top}
    &=
    \hat{\mathbf{U}}^{-}
    \left[
        \hat{\mathbf{D}}^{-}
        -
        \frac{1}{a}\mathbf{g}\mathbf{g}^{\top}
    \right]
    \bigl(\hat{\mathbf{U}}^{-}\bigr)^{\top}.
\end{aligned}
\end{equation}
Let
\begin{equation}
    \hat{\mathbf{D}}^{-} - \frac{1}{a}\mathbf{v}\mathbf{v}^{\top}
    =
    \tilde{\mathbf{U}}\tilde{\mathbf{D}}\tilde{\mathbf{U}}^{\top},
    \label{eq:rank_one_update}
\end{equation}
be the UD factorization of the bracketed matrix. The updated factors are then
\begin{align}
    \hat{\mathbf{U}}^{+} &= \hat{\mathbf{U}}^{-}\tilde{\mathbf{U}},
    \label{eq:U_measurement_update}\\
    \hat{\mathbf{D}}^{+} &= \tilde{\mathbf{D}}.
    \label{eq:D_measurement_update}
\end{align}
Carlson's rank-one update computes \eqref{eq:rank_one_update}--\eqref{eq:D_measurement_update} efficiently without explicitly forming the full covariance~\cite{Carlson1973_SquareRoot, Maybeck1979_SMEC}.

\begin{algorithm}[ht!]
\SetAlgoLined
\LinesNumbered
\KwIn{$\hat{\mathbf{U}}^{-},\,\hat{\mathbf{D}}^{-},\,\mathbf{H}\in\mathbb{R}^{1\times 2n},\,R$}
\KwOut{$\hat{\mathbf{U}}^{+},\,\hat{\mathbf{D}}^{+},\,\mathbf{K}$}
$\mathbf{f} \gets (\hat{\mathbf{U}}^{-})^{\top}\mathbf{H}^{\top}$\;
$\mathbf{g} \gets \hat{\mathbf{D}}^{-}\mathbf{f}$\;
$a_0 \gets R$\;
$\mathbf{b} \gets \mathbf{0}$\;
\For{$i=1$ \KwTo $2n$}{
    $a_i \gets a_{i-1} + f_i g_i$\;
    $D^{+}_{ii} \gets \dfrac{a_{i-1}}{a_i}\,D^{-}_{ii}$\;
    $b_i \gets g_i$\;
    $p_i \gets -\dfrac{f_i}{a_{i-1}}$\;
    \For{$j=1$ \KwTo $i-1$}{
        $U^{+}_{j i} \gets U^{-}_{j i} + p_i b_j$\;
        $b_j \gets b_j + g_i U^{-}_{j i}$\;
    }
}
$\mathbf{K} \gets \mathbf{b}/a_{2n}$\;
\caption{Carlson rank-one UD update for a scalar measurement~\cite{Carlson1973_SquareRoot, Maybeck1979_SMEC}.}
\label{alg:rank_one_update}
\end{algorithm}

\subsubsection{State and Covariance Update}

Given the Kalman gain $\mathbf{K}$ from Algorithm~\ref{alg:rank_one_update}, the augmented state estimate is updated as
\begin{equation}
    \hat{\mathbf{x}}_{k|k_m}^{\mathrm{sc}}
    =
    \hat{\mathbf{x}}_{k|k_{m-1}}^{\mathrm{sc}}
    +
    \mathbf{K}
    \left(
        z_{k,m}
        -
        h_{k,m}\!\left(\hat{\mathbf{x}}_{k|k_{m-1}}^{\mathrm{sc}}\right)
    \right).
\end{equation}

After processing all $M_k$ measurements at epoch $k$, the resulting augmented covariance can be expressed in block form as
\begin{equation}
    \hat{\mathbf{P}}_{k|k}^{\mathrm{sc}}
    =
    \hat{\mathbf{U}}^{+}\hat{\mathbf{D}}^{+}\bigl(\hat{\mathbf{U}}^{+}\bigr)^{\top}
    =
    \begin{bmatrix}
        \hat{\mathbf{P}}_{k|k} & \hat{\mathbf{P}}_{k,k-1|k} \\
        \hat{\mathbf{P}}_{k,k-1|k}^{\top} & \hat{\mathbf{P}}_{k-1|k}
    \end{bmatrix},
    \label{eq:augmented_covariance_update}
\end{equation}
where the UD factors have the block structure
\begin{equation}
    \hat{\mathbf{U}}^{+}
    =
    \begin{bmatrix}
        \hat{\mathbf{U}}^{+}_{11} & \hat{\mathbf{U}}^{+}_{12} \\
        \mathbf{0} & \hat{\mathbf{U}}^{+}_{22}
    \end{bmatrix},
    \qquad
    \hat{\mathbf{D}}^{+}
    =
    \begin{bmatrix}
        \hat{\mathbf{D}}^{+}_{11} & \mathbf{0} \\
        \mathbf{0} & \hat{\mathbf{D}}^{+}_{22}
    \end{bmatrix}.
\end{equation}

\subsubsection{Extracting the Non-augmented Covariance}

To recover the UD factors of the non-augmented covariance $\hat{\mathbf{P}}_{k|k}$, note from \eqref{eq:augmented_covariance_update} that
\begin{equation}
    \hat{\mathbf{P}}_{k|k}
    =
    \hat{\mathbf{U}}^{+}_{11}\hat{\mathbf{D}}^{+}_{11}\bigl(\hat{\mathbf{U}}^{+}_{11}\bigr)^{\top}
    +
    \hat{\mathbf{U}}^{+}_{12}\hat{\mathbf{D}}^{+}_{22}\bigl(\hat{\mathbf{U}}^{+}_{12}\bigr)^{\top}.
    \label{eq:rank_k_update}
\end{equation}
Equation~\eqref{eq:rank_k_update} is a rank-$n$ update of the top-left block and can be implemented by repeated rank-one UD updates (Agee--Turner) using each column of $\hat{\mathbf{U}}^{+}_{12}$ together with the corresponding diagonal element of $\hat{\mathbf{D}}^{+}_{22}$~\cite{Agee1972_TriangularDyad}, or by simply running Algorithm \ref{alg:udu_factorization} on computed $\hat{\mathbf{P}}_{k|k}$ directly.

%% file: sections/3_4_reducing_computation.tex
\subsection{Reducing Computation Cost for the Measurement Update}
\label{sec:computation_reduction}

At epoch $k$, suppose we have $m_1$ measurements that depend on both the current and cloned states (e.g., TDCP) and $m_2$ measurements that depend only on the current state (e.g., pseudorange), with $M_k = m_1 + m_2$. 
This subsection shows that, by ordering the sequential measurement updates appropriately, the computational cost of the UD measurement update can be reduced without changing the filtering result.

We first process the $m_1$ delayed-state measurements. After these updates, we obtain UD factors $(\hat{\mathbf{U}}^{+},\hat{\mathbf{D}}^{+})$ for the augmented covariance as described in Section~\ref{sec:meas_update}. 
We then process the remaining $m_2$ current-state-only measurements. For such measurements, $\mathbf{H}_{k,m}^{(2)}=\mathbf{0}$, so the auxiliary vectors in Algorithm~\ref{alg:rank_one_update} $\mathbf{f}$ and $\mathbf{g}$ inherit a block structure that does not require the bottom-right UD factor $\hat{\mathbf{U}}^{-}_{22}$:
\begin{equation}
\begin{aligned}
    \mathbf{f}
    &=
    (\hat{\mathbf{U}}^{-})^{\top}\mathbf{H}_{k,m}^{\top}
    =
    \begin{bmatrix}
        (\hat{\mathbf{U}}^{-}_{11})^{\top}\,(\mathbf{H}^{(1)}_{k,m})^{\top} \\
        (\hat{\mathbf{U}}^{-}_{12})^{\top}\,(\mathbf{H}^{(1)}_{k,m})^{\top}
    \end{bmatrix},
    \qquad
    \mathbf{g}
    =
    \hat{\mathbf{D}}^{-}\mathbf{f}
    =
    \begin{bmatrix}
        \hat{\mathbf{D}}^{-}_{11}(\hat{\mathbf{U}}^{-}_{11})^{\top}(\mathbf{H}^{(1)}_{k,m})^{\top} \\
        \hat{\mathbf{D}}^{-}_{22}(\hat{\mathbf{U}}^{-}_{12})^{\top}(\mathbf{H}^{(1)}_{k,m})^{\top}
    \end{bmatrix}, \\
    \hat{\mathbf{U}}^{-}
    &=
    \begin{bmatrix}
        \hat{\mathbf{U}}^{-}_{11} & \hat{\mathbf{U}}^{-}_{12} \\
        \mathbf{0} & \hat{\mathbf{U}}^{-}_{22}
    \end{bmatrix},
    \qquad
    \hat{\mathbf{D}}^{-}
    =
    \begin{bmatrix}
        \hat{\mathbf{D}}^{-}_{11} & \mathbf{0} \\
        \mathbf{0} & \hat{\mathbf{D}}^{-}_{22}
    \end{bmatrix}.
\end{aligned}
\label{eq:f_v_structure}
\end{equation}
Importantly, the covariance update equation \eqref{eq:rank_k_update} does not depend on $\Uhatplus_{22}$, and from \eqref{eq:f_v_structure}, the matrix $\Uhatminus_{22}$ (=$\Uhatplus_{22}$ from the previous measurement) is not involved in the computation of $\mathbf{f}$ and $\mathbf{g}$, which is used to compute $\Uhatplus_{11}$, $\Uhatplus_{12}$, $\Dhatplus_{11}$, and $\Dhatplus_{22}$ in Algorithm \ref{alg:rank_one_update}. In addition, the Kalman gain for the first half of the augmented state vector (i.e., the current state) can be computed without using the second half of the Kalman gain vector, which depends on $b_j, (j = n+1, \ldots, 2n)$ in Algorithm \ref{alg:rank_one_update}.

Therefore, when (1) processing the measurements that depend only on the current state, and (2) the remaining measurements at the same timestep do not depend on the past states, we can skip the computation of $\Uhatplus_{22}$ and $b_j (j > n)$ in Algorithm \ref{alg:rank_one_update}, by terminating the second loop (line 9) at $j = n$ instead of $j = k-1$ when $k > n + 1$. Note that we still need to run the first loop (line 4) from $k = 1$ to $2n$.
It is important that when (2) is not satisfied, we need to compute $\Uhatplus_{22}$ and $b_j (j > n)$ since they are needed for the future measurement update to process the measurements that depend on the past states. Thus, to reduce the computational cost, we should first process all the measurements that depend on both the current and past states, and then process the measurements that depend only on the current state.

When delayed-state measurements remain to be processed (either later within the same epoch, or because the implementation maintains the full augmented factors for subsequent smoothing), the full update must be retained; i.e., $\hat{\mathbf{U}}^{+}_{22}$ and $b_j$ for $j>n$ cannot be skipped. 
Consequently, to minimize computational cost while preserving correctness, we recommend processing all measurements that depend on both the current and cloned states first (e.g., TDCP), followed by measurements that depend only on the current state (e.g., pseudorange).

Note that if we wish to apply a fixed-interval smoother after filtering, the full augmented-state factors and state estimates must be retained for all measurements, since the smoothing recursion (Section~\ref{sec:smoothing}) requires the complete augmented covariance structure, including the cloned-state components.

%% file: sections/3_5_smoothing.tex
\subsection{Fixed-Interval Smoothing}
\label{sec:smoothing}
For post-processing applications, estimation accuracy is improved by applying a fixed-interval smoother. 
In standard models where no delayed-state measurements occur, the standard Rauch-Tung-Striebel (RTS) smoother~\cite{Rauch1965RTS} provides optimal estimates of the past states $\hat{\mathbf{x}}_{k|N}$ and covariances $\hat{\mathbf{P}}_{k|N}$ given all measurements up to time $N > k$.
In RTS smoother, the smoothed estimates are recursively computed backwards from $k=N$ to $k=1$:
\begin{align}
    \hat{\mathbf{x}}_{k | N} &= \hat{\mathbf{x}}_{k | k} + \mathbf{A}_k \left( \hat{\mathbf{x}}_{k+1 | N} - \hat{\mathbf{x}}_{k+1 | k} \right) \\
    \hat{\mathbf{P}}_{k | N} &= \hat{\mathbf{P}}_{k | k} + \mathbf{A}_k \left( \hat{\mathbf{P}}_{k+1 | N} - \hat{\mathbf{P}}_{k+1 | k} \right) \mathbf{A}_k^{\top}
\end{align}
where the standard smoother gain $\mathbf{A}_k$ is defined using the predicted covariance inverse:
\begin{equation}
    \mathbf{A}_k = \hat{\mathbf{P}}_{k | k} \mathbf{\Phi}_{k+1, k}^{\top} \hat{\mathbf{P}}_{k+1 | k}^{-1}
\end{equation}
Here, the standard RTS formulation relies on the fundamental Markov assumption of the standard state-space model:
\begin{equation}
    p(\mathbf{x}_k \mid \mathbf{x}_{k+1}, \mathbf{Z}_{1:N}) = p(\mathbf{x}_k \mid \mathbf{x}_{k+1}, \mathbf{Z}_{1:k})
\end{equation}
This assumption states that once the future state $\mathbf{x}_{k+1}$ is known, future measurements $\mathbf{Z}_{k+1:N}$ provide no additional information about the current state $\mathbf{x}_k$. In standard systems, the only link between $\mathbf{x}_k$ and $\mathbf{x}_{k+1}$ is the process model $\mathbf{\Phi}$, so conditioning on $\mathbf{x}_{k+1}$ effectively "blocks" the information flow from future measurements.

However, this assumption is violated in the presence of delayed-state measurements. A TDCP measurement $\mathbf{z}_{k+1}$ is a function of both the current and the previous state:
\begin{equation}
    \mathbf{z}_{k+1} = h_1(\mathbf{x}_{k+1}) - h_2(\mathbf{x}_k) + \mathbf{v}_{k+1}
\end{equation}
Because of this functional dependence, $\mathbf{z}_{k+1}$ provides direct information about $\mathbf{x}_k$ that is not captured by $\mathbf{x}_{k+1}$ alone. Consequently, the standard RTS gain $\mathbf{A}_k$, which accounts only for the process correlation via $\mathbf{\Phi}_{k+1, k}$, fails to capture the posterior correlation introduced by the measurement itself.

To correctly smooth the delayed state, we must derive the update from the joint posterior distribution obtained by the Stochastic Cloning (SC) filter, which naturally incorporates both process and measurement information. The joint estimate of the states at time step $k+1$ and $k$ is given by:
\begin{equation}
    P(\mathbf{x}_{k+1}, \mathbf{x}_{k} | \mathbf{Z}_{1:k+1}) = \mathcal{N} \left(
    \begin{bmatrix}
    \hat{\mathbf{x}}_{k+1 | k+1} \\
    \hat{\mathbf{x}}_{k| k+1}
    \end{bmatrix},
    \begin{bmatrix}
    \hat{\mathbf{P}}_{k+1 | k+1} & \hat{\mathbf{P}}_{k+1, k | k+1} \\
    \hat{\mathbf{P}}_{k+1, k | k+1}^{\top} & \hat{\mathbf{P}}_{k | k+1}
    \end{bmatrix}
    \right)
\end{equation}
Using the properties of conditional multivariate Gaussian distributions \cite{Barfoot2024_StateEstimationRobotics}, the conditional mean of $\mathbf{x}_{k}$ given $\mathbf{x}_{k+1}$ is:
\begin{equation}
    E[\mathbf{x}_{k} | \mathbf{x}_{k+1}, \mathbf{Z}_{1:k+1}] = \hat{\mathbf{x}}_{k| k+1} + \hat{\mathbf{P}}_{k+1, k | k+1}^{\top} \hat{\mathbf{P}}_{k+1 | k+1}^{-1} \left( \mathbf{x}_{k+1} - \hat{\mathbf{x}}_{k+1 | k+1} \right)
\end{equation}
The smoothed estimate of the delayed state is derived as:
\begin{equation}
\begin{aligned}
    \hat{\mathbf{x}}_{k | N} &= E \left[ E[\mathbf{x}_{k} | \mathbf{x}_{k+1}, \mathbf{Z}_{1:k+1}] \mid \mathbf{Z}_{1:N} \right] \\ 
    &= \hat{\mathbf{x}}_{k| k+1} + \mathbf{J}_k \left( \hat{\mathbf{x}}_{k+1 | N} - \hat{\mathbf{x}}_{k+1 | k+1} \right)
\end{aligned}
\end{equation}
Here, the smoothing gain $\mathbf{J}_k$ uses the posterior cross-covariance $\hat{\mathbf{P}}_{k+1, k | k+1}$, which explicitly captures the measurement-induced correlation:
\begin{equation}
    \mathbf{J}_k = \hat{\mathbf{P}}_{k+1, k | k+1}^{\top} \hat{\mathbf{P}}_{k+1 | k+1}^{-1}
\end{equation}
Note that we can compute $\mathbf{J}_k$ using the stored UD factors $(\mathbf{U}, \mathbf{D})$ from the forward filter without explicit matrix inversion. We solve the linear system:
\begin{equation}
    \left( \Uhat_{k+1|k+1} \Dhat_{k+1|k+1} \UhatT_{k+1|k+1} \right) \mathbf{J}_k^{\top} = \Uhatplus_{12} \Dhatplus_{22} \UhatplusT_{22}
\end{equation}
The efficient solution is obtained via substitution:
\begin{enumerate}
    \item Backward Substitution: Solve $\Uhat_{k+1|k+1} \mathbf{Y} = \Uhatplus_{12}$.
    \item Diagonal Scaling: Compute $\mathbf{Z}_{ij} = \mathbf{Y}_{ij} \cdot \Dhatplus_{22}(j) / \Dhat_{k+1|k+1}(i)$.
    \item Forward Substitution: Solve $\UhatT_{k+1|k+1} \mathbf{J}_k^{\top} = \mathbf{Z} \UhatplusT_{22}$.
\end{enumerate}
Finally, the smoothed covariance is updated as:
\begin{equation}
    \hat{\mathbf{P}}_{k | N} = \hat{\mathbf{P}}_{k | k+1} + \mathbf{J}_k \left( \hat{\mathbf{P}}_{k+1 | N} - \hat{\mathbf{P}}_{k+1 | k+1} \right) \mathbf{J}_k^{\top}
\end{equation}

\begin{figure}[ht!]
    \centering
    \includegraphics[width=\linewidth]{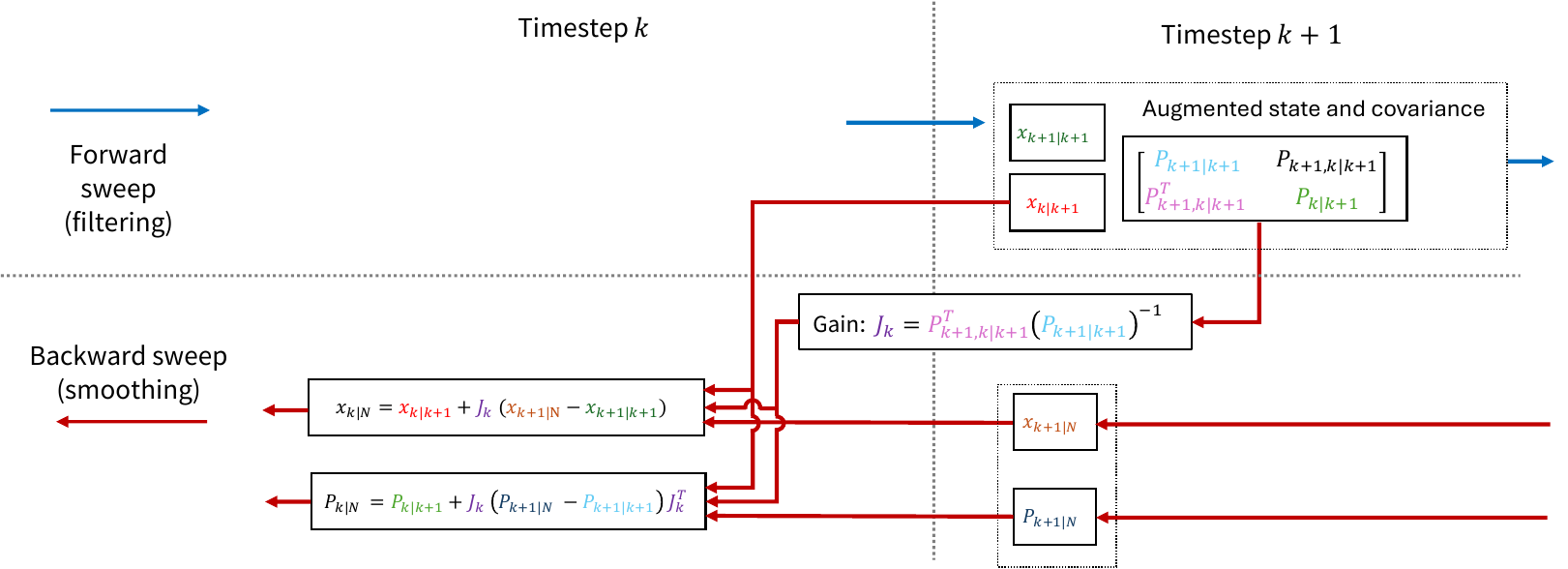}
    \caption{Relationship between the estimated states and covariances in backward smoothing}
    \label{fig:sc_smoother_diagram}
\end{figure}

The relationship between the estimated states and covariances in the forward and backward passes of the SC smoother is illustrated in Figure \ref{fig:sc_smoother_diagram}.

%% file: sections/4_pipeline.tex
\section{Estimation Pipeline}
\label{sec:pipeline}
In this section, we describe the overall estimation pipeline that integrates the delayed-state UD filter and smoother for lunar GNSS-based orbit and clock estimation. The pipeline consists of several key components: measurement preprocessing, state propagation, measurement update using the UD  filter, and post-processing with the smoother.

\subsection{Filtering}
\label{sec:pipeline_filtering}
\subsubsection{Time Update}
\label{sec:time_update_pipeline}
Given the state estimate $\hat{x}_{k-1|k-1}$ and covariance $P_{k-1|k-1}$ at time step $k-1$, the time update step propagates the state and covariance to the next time step $k$ as described in Section \ref{sec:time_update}. The dynamics model used for the state vector inside the filter is given by
\begin{align}
    f^{\text{dyn}}(t, \mathbf{x}) 
    &= \frac{d}{dt} \mathbf{x} \\
    &= \frac{d}{dt} \begin{bmatrix}
    \mathbf{r}_{k}&
    \mathbf{v}_{k} &
    c \delta t_{k} &
    c \delta \dot{t}_{k} &
    c \delta \ddot{t}_k &
    \gamma_{k}
    \end{bmatrix}^{\top} \\
    &= \begin{bmatrix}
    \mathbf{v}_k &
    \mathbf{a}(\mathbf{r}_k, \mathbf{v}_k, \gamma_k, t_k) &
    c \delta \dot{t}_k &
    c \delta \ddot{t}_k + c \dot{\delta} t_{rel}(r_k, v_k) &
    0 &
    0 
    \end{bmatrix}^{\top} 
\end{align}
For the orbital dynamics model, we include the high-order gravity terms from the Moon, third-body perturbations from the Earth and the Sun, as well as solar radiation pressure effects. \cite{Mina2025LCRNS} have shown that including the spherical harmonic gravity terms up to degree and order 18 is sufficient to achieve the acceleration fidelity of $3 \times 10^{-12}$ km/s$^2$.

The derivatives needed for the state transition matrix are given by
\begin{align}
    \frac{\partial f^{\text{dyn}}}{\partial \mathbf{x}} = F_k
    & = 
    \begin{bmatrix}
    \mathbf{0}_{3\times3} & \mathbf{I}_{3\times3} & \mathbf{0}_{3\times1} & \mathbf{0}_{3\times1} & \mathbf{0}_{3\times1} & \mathbf{0}_{3\times1} \\
    \frac{\partial \mathbf{a}}{\partial \mathbf{r}} & 
    \frac{\partial \mathbf{a}}{\partial \mathbf{v}}
    & \mathbf{0}_{3\times1} & \mathbf{0}_{3\times1} & \mathbf{0}_{3\times1} & \frac{\partial \mathbf{a}}{\partial \gamma} \\
    \mathbf{0}_{1\times3} & \mathbf{0}_{1\times3} & 0 & 1 & 0 & 0\\
    \left( \frac{\partial (c \dot{\delta} t_{rel})}{\partial \mathbf{r}} \right)^{\top} & 
    \left( \frac{\partial (c \dot{\delta} t_{rel})}{\partial \mathbf{v}} \right)^{\top} & 0 & 0 & 0 & 0 \\ 
    \mathbf{0}_{1\times3} & \mathbf{0}_{1\times3} & 0 & 0 & 1 & 0\\
    \mathbf{0}_{1\times3} & \mathbf{0}_{1\times3} & 0 & 0 & 0 & 0 \\
    \mathbf{0}_{1\times3} & \mathbf{0}_{1\times3} & 0 & 0 & 0 & 0 \\
    \end{bmatrix}
\end{align}
where the partial derivatives associated with relativistic time delays are given by
\begin{align}
    \frac{\partial (c \dot{\delta} t_{rel})}{\partial \mathbf{r}} &= -\frac{\mu_M}{2c}\cdot \frac{\mathbf{r}}{\|\mathbf{r}\|^3} \in \R^{3}\\
    \frac{\partial (c \dot{\delta} t_{rel})}{\partial \mathbf{v}} &= \frac{\mathbf{v}}{c} \in \R^{3}
\end{align}
The readers are referred to \cite{montenbruck2013satellite} for other partial derivatives. Process noise for the position and velocity state is modeled as \cite{Carpenter2025_NavFilterBestPractices}:
\begin{align}
    Q_{pos, vel} = \begin{bmatrix}
    \frac{1}{3} q_a \Delta t^3 \mathbf{I}_{3\times3} 
    & \frac{1}{2} q_a \Delta t^2 \mathbf{I}_{3\times3} \\
    \frac{1}{2} q_a \Delta t^2 \mathbf{I}_{3\times3} 
    & q_a \Delta t \mathbf{I}_{3\times3}
    \end{bmatrix}
\end{align}
where $q_a$ is the spectral density of the acceleration process noise, and $\Delta t$ is the time step duration. The process noise for the clock states are modeled using equation \eqref{eq:clock_process_noise}.

\subsubsection{Measurement Preprocessing}
The first step in the pipeline involves preprocessing the raw GNSS measurements received by the lunar satellite. For time step $k$ this includes:
\begin{enumerate}
    \item Signal Acquisition and Tracking: Acquire and track the GNSS signals to extract raw pseudorange and carrier phase measurements. The detailed algorithms for weak GNSS signal acquisition and tracking are beyond the scope of this paper. Readers are referred to \cite{Capuano2016StandaloneGPS, Simone2023LuGREReceiver} for more information.
    \item Time frame conversion: Compute the current (reception time) GPS time from the local time of the lunar satellite using the current estimated bias from TCL and the time frame conversion equations \eqref{eq:time_transformation}.
    \item Light time correction: Iteratively solve for equation \eqref{eq:light_time_equation} to compute the transmission time and position of each GNSS satellite from the decoded satellite ephemeris.
    \item Altitude masking: For each GPS satellite $i$, compute the tangential altitude of the ray as follows. 
    \begin{equation}
        h_{\perp, k, i} = \left\| {\mathbf{r}_{tx} - \left( \frac{\mathbf{r}_{tx} ^{\top} \left(\mathbf{r}_{tx} - \mathbf{r} \right)}{\| \mathbf{r}_{tx} - \mathbf{r} \|^2} \right) (\mathbf{r}_{tx} - \mathbf{r})} \right\| - R_E
    \end{equation}
    where $R_E$ is the Earth's mean radius, $\mathbf{r}_{tx}$ is the position of the transmitting GNSS satellite in ECEF coordinates, and $\mathbf{r}$ is the position of the lunar satellite receiver in the ECEF coordinates. Remove all the measurements with tangential altitudes below a certain threshold to mitigate the impact of ionospheric delays.
    \item Compute the Shapiro time delay $c \Delta t^s$  using \eqref{eq:shapiro_delay} and subtract them from the raw measurements.
    \item Compute the receiver noise of the pseudorange and carrier phase measurements based on the observed carrier-to-noise ratio \CN values.
    \item Compute the ionosphere-free pseudorange and time-differenced carrier phase measurements using \eqref{eq:if_pseudorange} and \eqref{eq:tdcp}. To avoid the time-correlation between two consecutive time-differenced carrier phase measurements, we process the TDCP measurements every other time step (i.e., at $k = 0, 2, 4, \dots$).
    \item Compute the predicted measurements $\hat{z}_{k, i}$ and their residuals $y_{k, i}$ using the current state estimate $\hat{\mathbf{x}}_{k|k-1}$:
    \item Outlier rejection: Apply a statistical outlier rejection method (e.g., 3-sigma rule) to remove any remaining outliers in the preprocessed measurements.
    \begin{equation}
        \text{If } |z_{k, i} - \hat{z}_{k, i}| > 3 \sqrt {\mathbf{H}_{k, i} \hat{P}_{i, i} \mathbf{H}_{k, i}^T + R_{k, i}} \text{ then discard measurement } z_{k, i}
    \end{equation}
    For TDCP, we apply a dedicated outlier detection method based on the innovation sequence to identify and remove cycle slips, as follows:
    \begin{equation}
        \begin{cases}
        \text{Accept TDCP measurement} \ i \ \text{if} & |z_{k, i} - \hat{z}_{k, i}| < 2.5 S_{k,i}, \ \text{and} \ 3S_{k, i}< \lambda_L \\
        \text{Discard TDCP measurement} \ i \ \text{if} & \text{otherwise}
        \end{cases}
    \end{equation}
    where $S_{k, i} = \sqrt {\mathbf{H}_{k, i} \hat{P}_{i, i} \mathbf{H}_{k, i}^T + R_{k, i}} $.
    This criterion discards the TDCP measurements whenever the residual exceeds a 2.5-sigma bound, or exceeds the wavelength of the signal $\lambda_L$ (=19 cm for L1 and 24 cm for L5). It also discards all measurements where the 3-sigma bound itself exceeds the wavelength, which indicates insufficient state knowledge and/or measurement noise to reliably detect cycle slips.
    For this work, we only discard the TDCP measurements with detected cycle slips, and do not attempt to repair them.
\end{enumerate}   

\subsubsection{Measurement Update}
\label{sec:pipeline_measurement_update}
The measurement updates are performed using the UD filter as described in Section \ref{sec:meas_update}. Assuming a dual-frequency receiver, we process two measurements: the ionosphere-free pseudorange and the TDCP for L1 signals. 
The measurement Jacobians for each measurement type are given by:
\begin{align}
    \mathbf{H}_{i}^{\rho} &= 
    \begin{bmatrix}
        \mathbf{H}_{i} & 0_{1 \times 10}  
    \end{bmatrix} \in \R^{1 \times 20} \\
    \mathbf{H}_{k, i}^{tdcp} &= \begin{bmatrix}
        \mathbf{H}_{k, i}  & \mathbf{H}_{k-1, i} 
    \end{bmatrix} \in \R^{1 \times 20} \\
    \mathbf{H}_{k, i} &= \begin{bmatrix}
    -\frac{(\mathbf{r}_{k} - \mathbf{r}_{tx, k, i})^{\top}}{\|\mathbf{r}_{k} - \mathbf{r}_{tx, k, i}\|} & \mathbf{0}_{1\times3} & 1 & 0 & 0 & 0
    \end{bmatrix} \in \R^{1 \times 10}
\end{align}
and the measurement noise variances are modeled as 
\begin{align}
    R_{k, i}^{\rho} &= \alpha_{L1}^2 \sigma_{\rho, k, i, L1}^2 + \alpha_{L5}^2 \sigma_{\rho, k, i, L5}^2 + \sigma_{\text{URE}}^2  \\
    R_{k, i}^{tdcp} &= \sigma_{\phi, k, i, L1}^2 + \sigma_{\phi, k-1, i, L1}^2  + \sigma_{\Delta \text{URE}}^2 
\end{align}
where $\sigma_{\rho, k, i, l}$, and $\sigma_{\phi, k, i, l}^2$ are the thermal noise in the Delay Lock Loop (DLL) and Phase Lock Loop (PLL) for frequency band $l$ at time step $k$ for satellite $i$, respectively, and $\sigma_{\text{URE}}^2$ and $\sigma_{\Delta \text{URE}}^2$ are the variances of the unmodeled residual errors for pseudorange and TDCP measurements, respectively, mainly due to ephemeris errors for pseudorange and time-differenced ephemeris and ionospheric errors for TDCP. 
While inflating the noise variances for these biases are not optimal from a Bayesian perspective, it provides a simple and effective way to prevent overconfident state covariance estimates without explicitly estimating them, which leads to increased computational complexity and potential observability issues, since it is challenging to distinguish biases, signal direction orbit errors, and clock errors under poor geometry conditions.

\subsection{Post-Processing}
\label{sec:pipeline_post_processing}
Finally, at certain time intervals, we apply the delayed-state smoother as described in Section \ref{sec:smoothing} to refine the state estimates using all available measurements up to the final time step $N$. 
Often, further accuracy improvements can be achieved by iteratively applying the filter and smoother multiple times. 

Given the refined state estimates, the orbit and clock are propagated forward to predict future states, and are converted to ephemeris and clock products~\cite{Iiyama2025_EphemerisAlmanacDesign, Salgueiro2025_NovelNavMessageLCNS} for broadcasting.
Note that since the clock biases are estimated with respect to TCL in the filter and smoother, they need to be converted to LT if the ephemeris and clock products are defined in LT.

%% file: sections/5_simulation_setup.tex
\section{Simulation Setup}
\label{sec:simulation_setup}
In this section, we describe the simulation setups used to evaluate the proposed lunar GNSS-based orbit and clock estimation method. 

\subsection{Lunar Orbit}
\label{sec:lunar_orbit}
\begin{figure}[ht!]
    \centering
    \includegraphics[trim={5mm, 5mm, 5mm, 5mm}, width=0.5\textwidth]{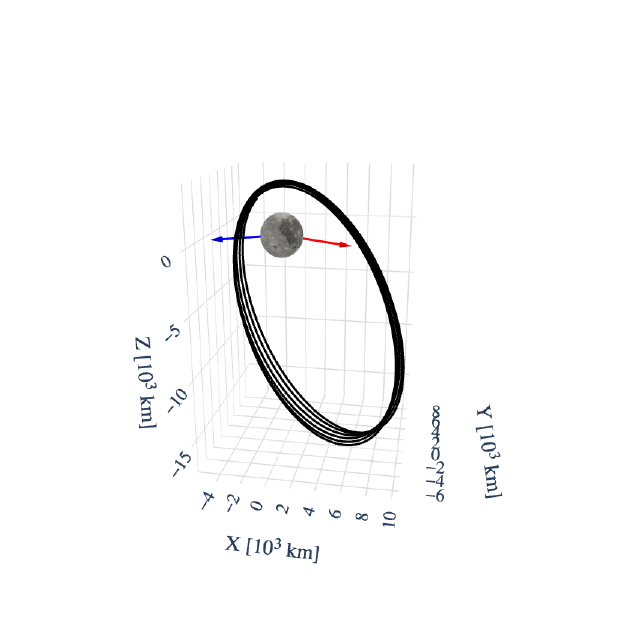}
    \caption{The simulated LCRNS (LDN-1) satellite orbit in LCRF (Moon-centered inertial frame) over 6 orbits (180 hours). The direction to Earth and Sun at the initial epoch is shown as blue and red arrows, respectively.}
    \label{fig:lunar_orbit}
\end{figure}

We used the initial conditions provided by \cite{LCRNS2025} to simulate LCRNS satellites (LDN-1). Initial epoch of the simulation was set to March 1, 2025, 12:00:00 UTC, since the precise orbits, ephemeris, and the parameters needed for the ionospheric and plasmaspheric delay simulations were not available for the date in the reference (March 1, 2027, 00:00:00 UTC). 
To convert the initial conditions, the initial position and velocity vectors of the LCRNS satellite given at \cite{LCRNS2025} were first converted to orbital elements at OP-frame (Orbital Plane frame) defined in \cite{Ely2005FrozenLunarOrbits}, and then the same initial orbital elements were used to generate the position and velocity vectors at the new initial epoch (March 1, 2025, 12:00:00 UTC) in the OP-frame. 
Finally, the position and velocity vectors were transformed back to the LCRF for the orbit propagation. 
The initial orbital elements in the OP-frame are summarized in Table \ref{tab:lunar_orbit_elements}, and the propagated orbit is shown in Figure \ref{fig:lunar_orbit}.
The truth orbit was propagated using a lunar gravity model (up to degree and order 50), including third-body perturbations from the Earth and the Sun, as well as solar radiation pressure effects in a cannonball model with parameters $C_R=1.8$, satellite mass 850 kg, and 1.0 m$^2$ cross-sectional area, following \cite{LCRNS2025}. 
The simulation duration was set to 180 hours, corresponding to approximately 6 orbits.

\begin{table}[ht!]
    \centering
    \caption{The initial orbital elements of the LCRNS (LDN-1) satellite in the OP-frame at March 1, 2025, 12:00:00 UTC}
    \begin{tabular}{ l | c } \hline \hline
        Orbital element & Value \\ \hline
        Semi-major axis &  11315.93 km \\
        Eccentricity &  0.69198 \\
        Inclination &  61.208 deg \\
        Right ascension of ascending node &  116.90 deg \\
        Argument of periapsis &  85.21 deg \\
        True anomaly &  0.0 deg \\
        \hline \hline
    \end{tabular}
\label{tab:lunar_orbit_elements}
\end{table}

\subsection{GNSS Orbits and Ephemerides}
\label{sec:gnss_orbits}
\begin{figure}[ht!]
    \centering
    \includegraphics[trim={5mm, 5mm, 5mm, 5mm}, width=0.5\textwidth]{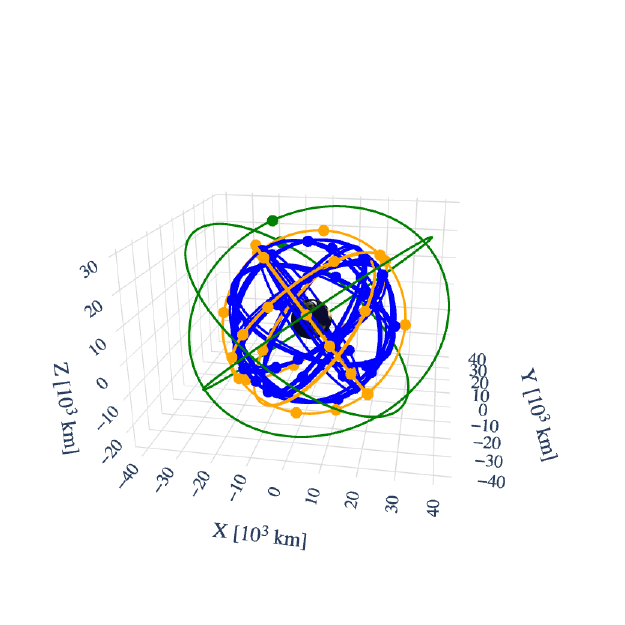}
    \caption{GNSS transmitter orbits in the J2000 frame generated from IGS final products (SP3). Blue, orange, and green markers denote GPS, Galileo, and QZSS satellites, respectively. Dots indicate satellite positions at the initial simulation epoch (March 1, 2025, 12:00:00~UTC).}
    \label{fig:gnss_orbits}
\end{figure}

We considered a multi-constellation set of GPS, Galileo, and Quasi-Zenith Satellite System (QZSS) satellites. Truth GNSS orbits were generated using the International GNSS Service (IGS) final orbit products~\cite{Johnston2017}. Specifically, final orbit products provide satellite center-of-mass positions, which were interpolated using cubic splines to obtain satellite positions at the desired epochs. Antenna phase center information from ANTEX files was then used to map center-of-mass positions to antenna reference point (ARP) positions.
The resulting GNSS constellation geometry at the simulation epoch is shown in Figure~\ref{fig:gnss_orbits}.

To emulate real-time navigation message performance, broadcast ephemerides were retrieved from daily multi-GNSS broadcast ephemeris files via Crustal Dynamics Data Information System (CDDIS). 
The broadcast ephemerides were used within the measurement models (Section~\ref{sec:gnss_navigation}) to introduce realistic ephemeris and clock modeling errors relative to the truth orbits. 

In addition, to remove the systematic bias common to all satellite clocks arising from differences in the time references used by the broadcast and precise clock products, the precise clock bias used as truth is adjusted as~\cite{Wang_2019}
\begin{equation}
\delta t^{\mathrm{pr}}_k \leftarrow \delta t^{\mathrm{pr}}_k
- \operatorname{median}\left( \delta t^{\mathrm{pr}} - \delta t^{\mathrm{br}} \right),
\end{equation}
where $\delta t^{\mathrm{br}}_{k}$ is the broadcast clock bias at time step $k$, and the median is computed over the entire simulation interval for each constellation.
For QZSS satellites, an analogous correction is applied to orbits on a per-satellite basis to remove the constant offset between the precise and broadcast ephemerides in the radial direction.

\subsection{Link Budget Analysis and Satellite Visibility}
The link budget analysis is performed to compute the carrier-to-noise density ratio (\CN) for the GNSS signals received by the lunar GNSS receivers. The \CN is calculated using the following equation~\cite{Konitzer2022LuGRE}:
\begin{equation}
\begin{aligned}
    (C/N_0)_{\text{dB-Hz}} &= P_{tx} + G_{tx} (\theta_{tx}, \varphi_{tx}) + G_{rx} (\theta_{rx}) -  20 \log_{10}\left(\frac{4 \pi d f}{c}\right) - L_{pol} - 10 \log_{10}(k_b T_{sys}) - R_{loss}
\end{aligned}
\end{equation}
where $P_{tx}$ and $G_{tx}$ is the transmit power and antenna gain of the GNSS satellites, $\theta_{tx}, \varphi_{tx}$ are the transmit antenna off-boresite angle and azimuth, $\theta_{rx}$ are the receiver antenna off-boresite angle and azimuth, $G_{rx}$ is the receiver antenna gain, $L_{path}$ is the free-space path loss, $L_{pol}$ is the polarization loss, $R_{loss}$ is the system implementation loss, $k_b = 1.38 \times 10^{-23}$ is the Boltzmann constant, and $T_{sys}$ is the system noise temperature.

\begin{figure}[ht!]
    \centering
    \includegraphics[width=\linewidth]{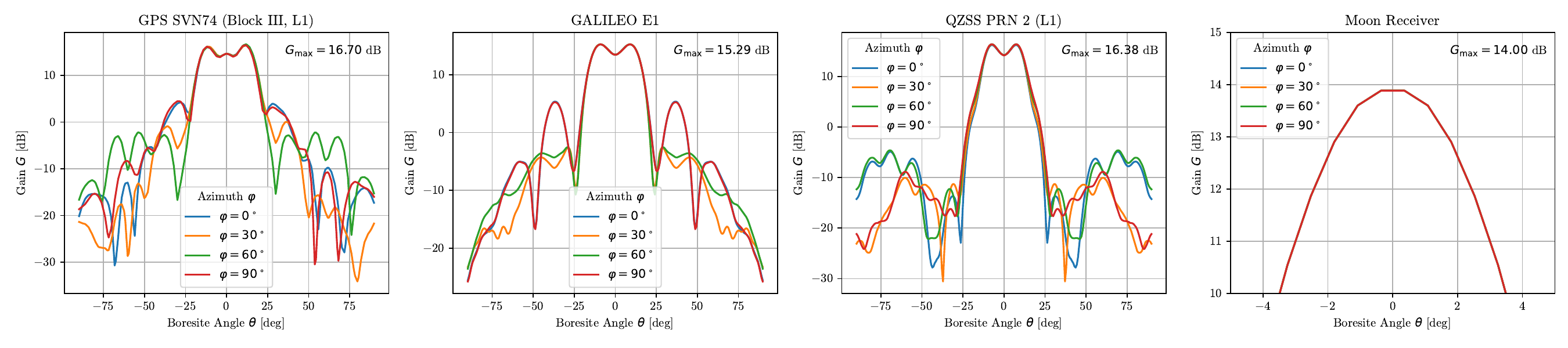}
    \caption{Antenna patterns of the selected GNSS satellites and the GNSS receiver onboard the satellite (From left: GPS SVN74 (Block III, L1), Galileo E1, QZS-2 (L1), Lunar GNSS receiver)}
    \label{fig:antenna_pattern}
\end{figure}

The transmitter power of the GNSS satellites $P_{tx}$ is summarized in Table \ref{tab:link_budget_params}, which is based on the values in \cite{Konitzer2022LuGRE} (from the ACE experiment and Magnetospheric Multiscale (MMS) Mission results) for GPS satellites, and \cite{Nakajima2024} for QZSS. 
We also add an additional 3 dBW to the Effective Isotropic Radiated Power (EIRP) values for the L5 signals.
Note that the value is not given for the Galileo satellites, since the Publications Office of the European Union~\cite{Menzione2024} publishes the EIRP values instead of the transmit power and the antenna gain separately. 
For the GNSS antenna patterns $G_{tx}$, we used the parameters from the NASA antenna characterization experiment~(ACE) study~\cite{donaldson2020characterization} for Block II-F,  the original data by Lockheed Martin for Block IIR and IIR-M satellites \cite{Marquis2015}, the United States Coast Guard Navigation Center~\cite{Fischer2022} for Block-III, Publications Office of the European Union~\cite{Menzione2024} for Galileo satellites, and the Cabinet Office of the Government of Japan for QZSS~\cite{QZSS2023}.
The receiver antenna gain $G_{rx}$ has a peak gain of 14 dBi and a half-power beamwidth of 12.2 degrees, taken from \cite{Delepaut2020GNSSLunar}.
The antenna patterns of the selected GNSS satellites and the GNSS receiver we assumed for the lunar users are shown in Figure \ref{fig:antenna_pattern}.

The computation of the angles $\theta_{tx}, \varphi_{tx}$ and $\theta_{rx}$ require the knowledge of the orientations of both the GNSS satellites and the lunar GNSS receiver.
The orientation of the GNSS satellites was always assumed to be in the "nominal yaw steering", where the antenna boresite is aligned with the spacecraft-Earth direction, and the solar panel rotation axis is orthogonal to the Earth-satellite-Sunplane~\cite{Montenbruck2015GNSS}. The off-nominal yaw steering models during the eclipse periods were not considered in this analysis, and is left for future work.
The orientation of the lunar GNSS receiver was assumed to be Earth-pointing, where the gimbaled antenna boresite is always aligned with the Moon-Earth direction.

\begin{table}[ht!]
    \centering
    \caption{The parameters used in the link budget analysis}
    \begin{tabular}{ l |l | c | c | c } \hline \hline
        Parameter & Refs & Symbol & Value & Units \\ \hline
        GPS transmit power (L1) & \cite{Konitzer2022LuGRE} & $P_{tx}$ & 17.3 (Block II-R), 18.8 (IIR-M), 16.2 (IIF) 18.8 (III) & dBW \\
        QZSS transmit power (L1) & \cite{Nakajima2024} & $P_{tx}$ & 14.1 & dBW \\
        Receiver peak antenna gain & \cite{Delepaut2020GNSSLunar} & $G_{rx}$ & 14.0 & dBi \\
        Receiver half-power beamwidth & \cite{Delepaut2020GNSSLunar} & $\theta_{3dB}$ & 12.2 & deg \\
        System noise temperature & \cite{Konitzer2022LuGRE} & $T_{sys}$ & 162.0 & K \\
        Polarization loss & \cite{Konitzer2022LuGRE} & $L_{pol}$ & 1.0 & dB \\
        Implementation loss & \cite{Konitzer2022LuGRE} & $R_{loss}$ & 0.9 & dB \\
        \hline \hline
    \end{tabular}
\label{tab:link_budget_params}
\end{table}

\subsection{Clock and GNSS Receiver Models}
\label{sec:clock_and_receiver_models}
The clock parameters are modeled based on typical space-grade oven-controlled crystal oscillators (OCXOs) to be used for LDN-1 satellite  \cite{Mina2025LCRNS}, as summarized in Table \ref{tab:clock_params}. 
Since the receiver parameters used in the receiver for the LuGRE mission was not publicly available, we set the parameters so that the noise profile will be close to the plots in \citep{Konitzer2022LuGRE}.
The receiver parameters used in the simulation are summarized in Table \ref{tab:gnss_receiver_params}.

\begin{figure}[ht!]
    \centering
    \includegraphics[width=0.8\textwidth]{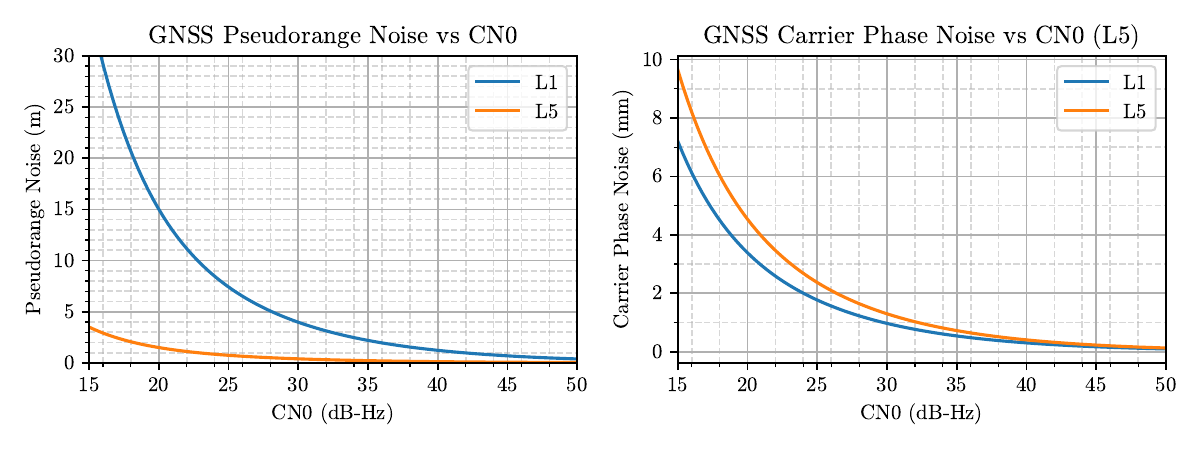}
    \caption{The relationship between the carrier-to-noise density ratio (\CN) and the pseudorange and carrier phase measurement noises for the lunar GNSS receiver used in the simulation.}
    \label{fig:measurement_noise_vs_cn0}
\end{figure}

Using the parameters given in Table \ref{tab:gnss_receiver_params}, and assuming thermal noise from the DLL and PLL are the dominant sources for the noise, the pseudorange and carrier phase measurement noises are computed as follows~\cite{kaplan2017understanding}:
\begin{align}
    \sigma_{\rho}^2 &= (c T_c)^2 \left[\frac{B_n}{2 C/N_0} \left( \frac{1}{B_{fe}T_c} \right) \left( 1 + \frac{1}{T_n (C/N_0)}  \right)  \right] \\
    \sigma_{\phi}^2 &= \left(\frac{\lambda_L}{2 \pi}\right)^2 \left[\frac{B_p}{2 C/N_0}  \left( 1 + \frac{1}{2T_p (C/N_0)}  \right)  \right]
\end{align}
where $\lambda_L$ is the wavelength of the GNSS signal, and $T_c$ is the code chip width of the GNSS signal. The relationship between the \CN and the measurement noises is shown in Figure \ref{fig:measurement_noise_vs_cn0}.

\begin{table}[ht!]
    \centering
    \caption{The power-spectral density parameters of the clock used in the simulation \cite{Mina2025LCRNS}}
    \begin{tabular}{ l | c } \hline \hline
        Parameter & Value \\ \hline 
        $q_1$ &  6.2299445014 $\times 10^{-13} \ s^{1/2}$ \\
        $q_2$ &  2.0129544799 $\times 10^{-14} \ s^{-1/2}$\\
        $q_3$ &  7.0118586804 $\times 10^{-28} \ s^{-3/2}$ \\
        \hline \hline
    \end{tabular}
\label{tab:clock_params}
\end{table}

\begin{table}[ht!]
    \centering
    \caption{The lunar GNSS receiver parameters used in the simulation}
    \begin{tabular}{ l | c | c | c } \hline \hline
        Parameter & Symbol & Value & Units \\ \hline
        DLL noise bandwidth & $B_{n}$ & 0.7 & Hz \\
        PLL noise bandwidth & $B_{p}$ & 1.0 & Hz \\
        Early-late correlator spacing & $D$ & 0.1 & chips \\
        DLL integration time & $T_n$ & 20 & ms \\
        PLL integration time & $T_p$ & 20 & ms \\
        Front-end bandwidth & $B_{fe}$ & 2 & MHz \\
        \hline \hline
    \end{tabular}
\label{tab:gnss_receiver_params}
\end{table}

\subsection{Ionospheric and Plasmapsheric Delays}
\label{sec:ionospheric_delays}

\begin{figure}[ht!]
    \centering
    \includegraphics[width=0.8\textwidth]{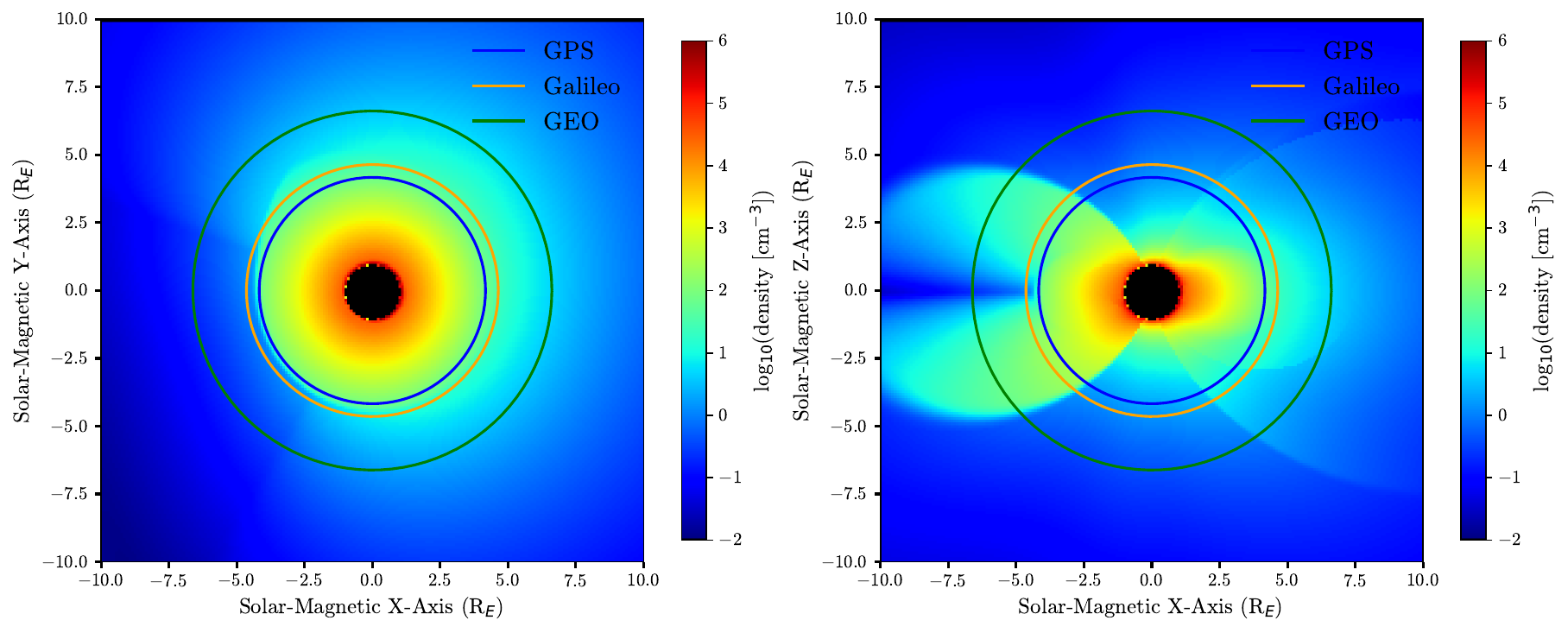}
    \caption{The simulated electron density profiles at March 1, 2025, 12:00:00 UTC using the IRI and GCPM models at the X-Y plane (left) and the X-Z plane (right) in 
    the solar magnetic coordinate system ($-X:$ toward the Sun, $Z:$ Earth's geomagnetic pole).
    The color bar indicates the electron density in units of electrons per cubic centimeter. The blue, orange, and green lines represent the orbit altitudes of the GPS (20,200 km), Galileo (23,222 km), and GEO (35,786 km), respectively.}
    \label{fig:iono_plasma_delay}
\end{figure}

The ionospheric and plasmaspheric delays are modeled by integrating the rays along the GNSS-to-lunar satellite. The path of the ray is governed by the refractive index (which can be computed from electron density) profile of the ionosphere and plasmasphere. The simulation uses the International Reference Ionosphere (IRI) model \cite{Bilitza2022IRIReview} for the ionosphere and the Global Core Plasma Model (GCPM) \cite{Gallagher2000GCPM} for the plasmasphere. 
Figure \ref{fig:iono_plasma_delay} shows the simulated electron density profiles at the equator and 45 degrees latitude at March 1, 2025, 12:00:00 UTC, which is the initial epoch of the simulation.
To model the electron densities under moderate solar and geomagnetic activity, we set the $K_p$ index is set to 3, and the 12-month running mean of the daily sunspot number (Rz12) to 50 for this plot and the rest of the paper.

By propagating the rays using the vector eikonal equations and applying the corerections proposed in our prior work~\cite{Iiyama2025Iono_iongnss, iiyama2025iono_arxiv}, we can simulate the bending of the GNSS signals as they propagate through the ionosphere and plasmasphere, which not only increases the distance traveled by the signals but also alters their propagation direction, changing the total electron content (TEC) along the LOS. 
However, since applying the corrections for all rays is computationally expensive, we only apply the corrections for signals with altitudes below $2000$ km, where the ray-bending effects are more pronounced. 
For signals with altitudes above $2000$ km, we assume straight-line rays for computational efficiency, and assume that the ionospheric and plasmaspheric delays are coming from the first-order delays only, which are proportional to the TEC along the LOS.
To further reduce the computational load, we compute the ionospheric and plasmaspheric delays at 120-second intervals and interpolate the delays for the measurement epochs at 1-second intervals.

%% file: sections/6_meas_error.tex
\section{Measurement Analysis}
\label{sec:meas_error_analysis}

\subsection{Satellite Availability}
\label{sec:satellite_availability}
\begin{figure}[ht!]
    \centering
    \includegraphics[width=0.95\textwidth]{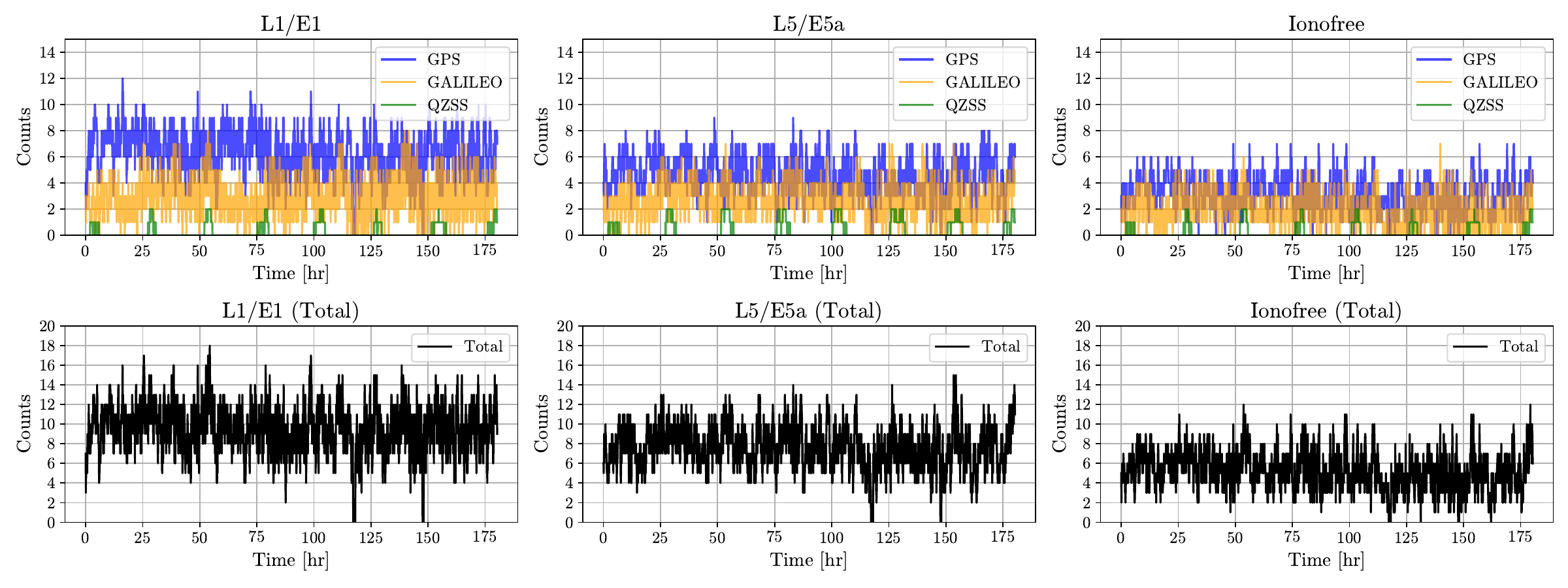}
    \caption{The number of tracked GNSS satellites over time for an acquisition and tracking threshold of \CN $>$ 18 dB-Hz for different GNSS constelations.}
    \label{fig:satellite_availability}
\end{figure}

Figure \ref{fig:satellite_availability} shows the satellite availability for a lunar GNSS receiver over the entire simulation period. A satellite is considered available if the receiver can track the satellite signal with a \CN greater than 18 dB-Hz. Note that \cite{Delepaut2020GNSSLunar} points out that higher \CN will be required to decode the almanac and ephemeris from the navigation messages. For this paper, we assume the ephemeris information is provided by uplink from the ground.

We see that the L1/E1 satellite has the most available satellites, followed by the L5/E5a satellites, since L5 signals are transmitted only from Block IIF and Block III satellites. The iono-free combination has the lowest number of available satellites, due to the need for tracking both L1 and L5 signals simultaneously.

Note that since the Earth's direction is nearly orthogonal to the orbital plane of the lunar satellite, we do not observe significant outages in satellite availability during the simulation period, due to lunar occultation of the GNSS satellites. In future works, we will investigate different lunar satellite orbits with different right ascension of ascending nodes (RAAN) to analyze the effect of lunar occultation on the estimation accuracy.

\subsection{Satellite Ephemeris Errors}
\label{sec:ephemeris_errors}
The broadcast ephemeris data are retrieved from the daily multi-GNSS Broadcast ephemeris files from CDDIS to simulate the real-time ephemeris errors. 
Assuming the precise ephemeris data from IGS (corrected for antenna phase center offsets) as the ground truth, we compute the ephemeris errors by comparing the broadcast ephemeris positions with the precise ephemeris positions for each GNSS satellite over the simulation period.

\begin{figure}[ht!]
    \centering
    \includegraphics[width=0.9\textwidth]{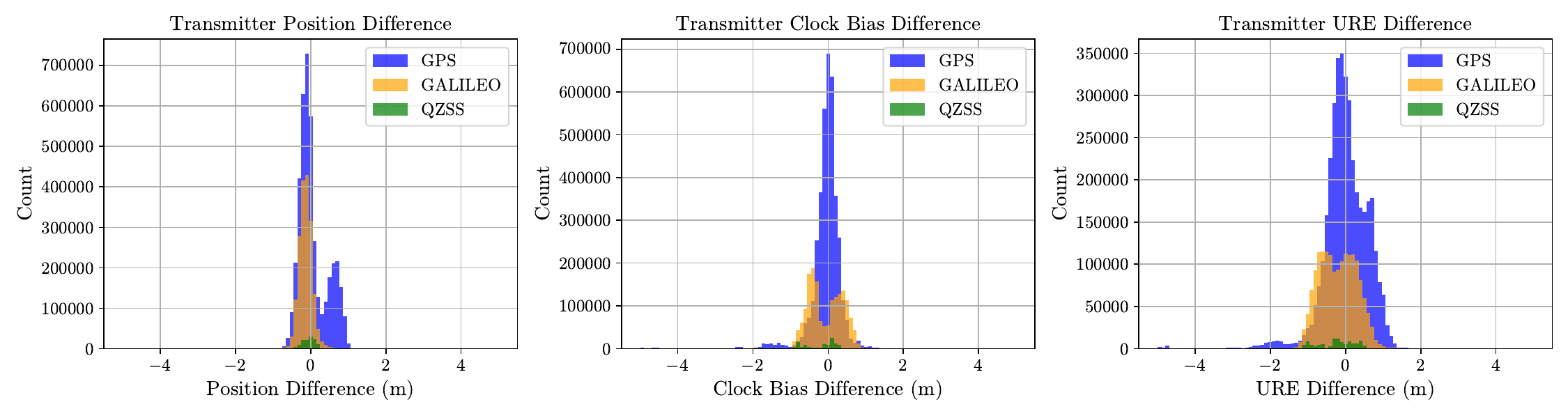}
    \caption{GNSS satellite ephemeris error histograms in the line-of-sight direction for different GNSS constellations}
    \label{fig:ephemeris_errors}
\end{figure}

The ephemeris error distributions for the GPS (L1/L5), Galileo (E1/E5a), and QZSS (L1/L5) satellites in directions along the GNSS-to-Lunar satellite line-of-sight is shown in Fig. \ref{fig:ephemeris_errors}. We see that most of the position and clock bias ephemeris errors are within $\pm 2$ m. The total user range error from the ephemeris are mostly within $\pm 3$ m, and has different mean biases for different GNSS constellations. The mean and variances of the ephemeris errors in LOS direction for different GNSS constellations are summarized in Table \ref{tab:ephemeris_error_stats}.

\begin{table}[ht!]
    \centering
    \caption{GNSS satellite ephemeris error statistics in the line-of-sight direction. Samples with absolute error greater than 10 m are excluded from the mean and standard deviation computation.}
    \begin{tabular}{c|c c|c c| c c}
        \hline \hline
        \multirow{2}{*}{GNSS Constellation} & \multicolumn{2}{c}{Position Error} & \multicolumn{2}{c}{Clock Bias Error} & \multicolumn{2}{c}{Total Range Error} \\
        \cline{2-7}
         & Mean (m) & Std. Dev. (m) & Mean (m) & Std. Dev. (m) & Mean (m) & Std. Dev. (m) \\    
        \hline
        GPS     &  0.054 & 0.372 & -0.089 & 0.772 & -0.037 & 0.886 \\
        Galileo & -0.135 & 0.177 & -0.078 & 0.648 & -0.215 & 0.679 \\
        QZSS    & -0.055 & 0.157 & -0.198 & 0.599 & -0.259 & 0.659 \\
        \hline \hline
    \end{tabular}
    \label{tab:ephemeris_error_stats}
\end{table}

\subsection{Receiver Noise}
\label{sec:receiver_noise}
\begin{figure}[ht!]
    \centering
    \includegraphics[width=0.9\textwidth]{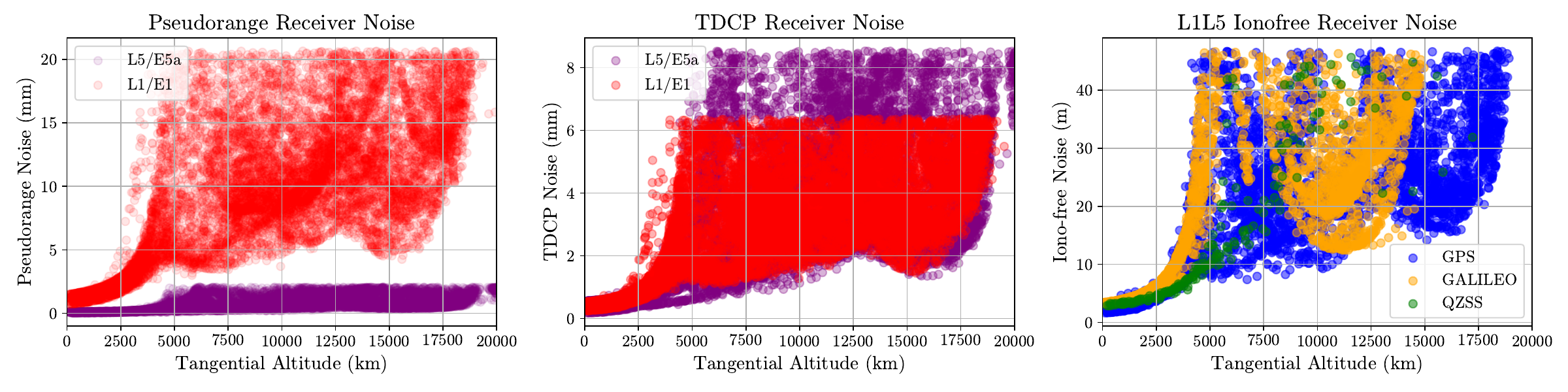}
    \caption{GNSS receiver noise vs tangential altitude for the L1 pseudorange (left), TDCP (center), and ionosphere-free combination (right), evaluated over a 180-hour simulation for an LCRNS satellite. The main lobe signals ($\leq$ 2000 km) have lower noise compared to the sidelobe signals with lower $C/N_0$.}
    \label{fig:receiver_noise}
\end{figure}

Figure \ref{fig:receiver_noise} shows the scatter plot of the pseudorange and carrier phase receiver noise errors for a lunar GNSS receiver, for different tangential altitudes of the ray from Earth. The points include all measurements sampled every 600 seconds over the entire simulation period. The receiver noise model described in Section \ref{sec:clock_and_receiver_models} is used to generate the measurement errors.
We observe that the receiver noise increases as the tangential altitude increases, due to the weaker signal strength from the GNSS satellites. The pseudorange receiver noise is also larger for L1 signals compared to L5 signals, due to the higher chipping rate of the L5 signals (10.23 Mcps) compared to L1 signals (1.023 Mcps). The TDCP noise is smaller for the L1 signals compared to the L5 signals, due to the wavelength difference (approximately 19 cm for L1 and 24 cm for L5). The iono-free combination pseudorange has larger noise compared to the L1 and L5 pseudorange measurements, due to the noise amplification from the linear combination.

\subsection{Ionospheric and Plasmaspheric Delays}
\label{sec:ionospheric_delays}
\begin{figure}[ht!]
    \centering
    \includegraphics[width=0.9\textwidth]{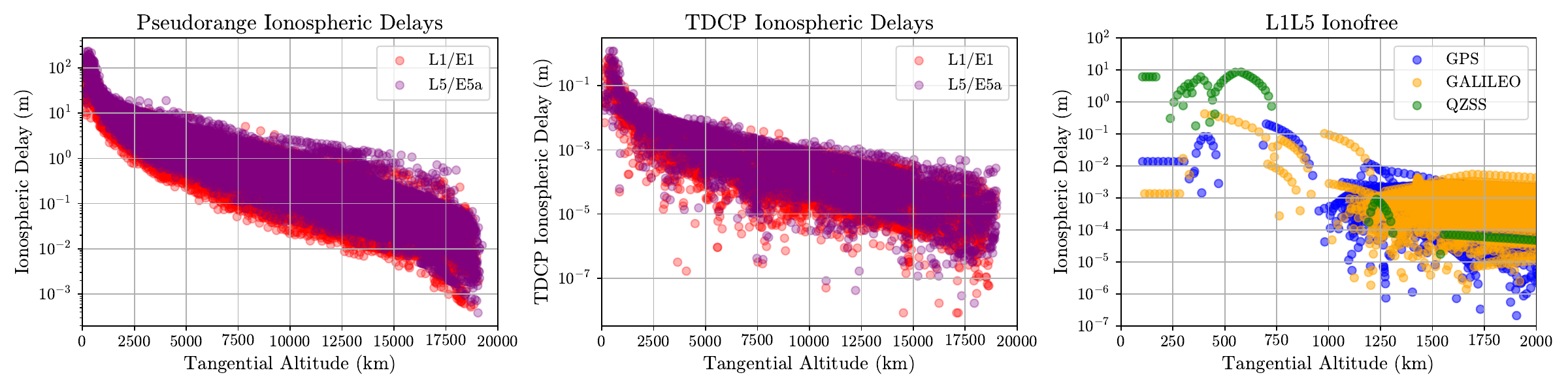}
    \caption{Ionospheric and plasmaspheric group delays vs tangential altitude for the L1 pseudorange (left), TDCP (center), and ionosphere-free combination (right), evaluated over a 180-hour simulation for an LCRNS satellite. For signal paths intersecting the ionosphere (tangential altitude $\leq 1000$ km), higher-order ionospheric effects and ray-bending delays, which are not fully eliminated by the ionosphere-free combination, remain observable (right).}
    \label{fig:iono_delay_scatter}
\end{figure}

Figure \ref{fig:iono_delay_scatter} shows the scatter plot of the ionospheric and plasmaspheric delays for pseudorange and carrier phase measurements for different tangential altitudes of the ray from Earth. The points include all measurements sampled every 600 seconds over the entire simulation period. We see that the ionospheric and plasmaspheric delays can reach more than 10 m for pseudorange measurements, and more than 100 mm for TDCP measurements, when the signals are passes through the ionosphere (tangential altitude less than 1000 km). The iono-free combination will remove most of the ionospheric and plasmaspheric delays, but some residual delays of up to 10 m can still be observed for low tangential altitudes (less than 1000 km), due to higher-order ionospheric effects and bending delays that are not removed by the first-order iono-free combination.

\subsection{Time Differenced Carrier Phase Measurement Errors}
\label{sec:tdcp_errors}
\begin{figure}[ht!]
    \centering
    \includegraphics[width=0.9\textwidth]{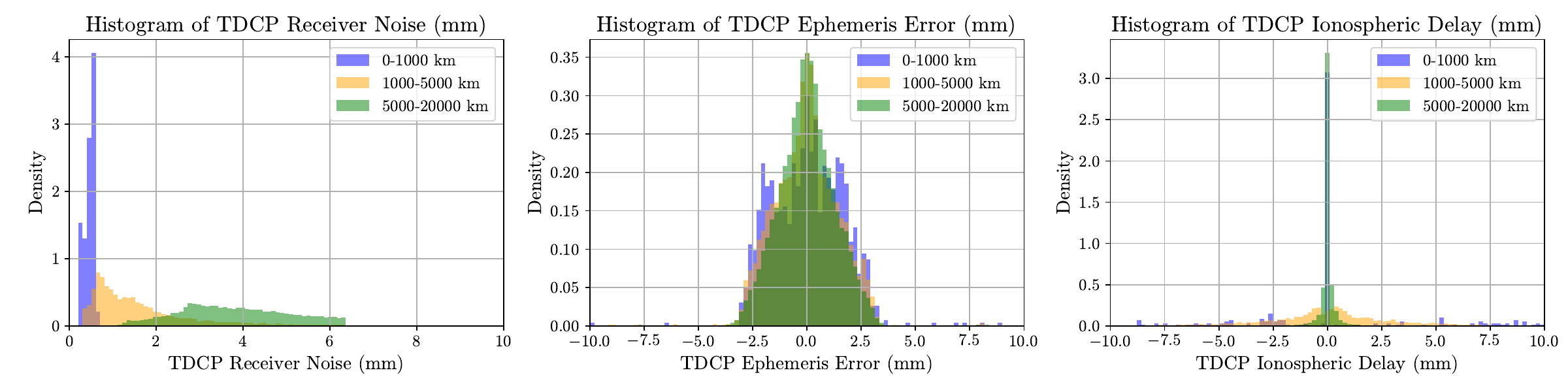}
    \caption{The distribution of the TDCP receiver noise, ephemeris errors, and ionospheric delays for different ranges of tangential altitudes.}
    \label{fig:tdcp_error_scatter}
\end{figure}

Figure~\ref{fig:tdcp_error_scatter} decomposes TDCP measurement errors into receiver thermal noise, broadcast ephemeris-induced error, and ionospheric/plasmaspheric residual error for three tangential-altitude regimes: low (0--1000~km), medium (1000--5000~km), and high (5000--20000~km). Samples are taken every 10~s over the simulation. The TDCP differencing interval is 1~s.

Table~\ref{tab:tdcp_error_stats} summarizes the mean and standard deviation of each error component, excluding outliers with absolute error greater than 10~m. The results highlight a key trade-off. At low and medium tangential altitudes, ionospheric/plasmaspheric residuals exhibit the largest dispersion (standard deviation $\approx$3--3.5~mm), exceeding the mean receiver-noise contribution (0.46~mm at low altitude and 1.63~mm at medium altitude). At high tangential altitudes, ionospheric residuals become small (standard deviation $\approx$0.34~mm), and TDCP error is dominated by receiver noise (mean $\approx$3.80~mm) due to weaker signals.

This behavior motivates tangential-altitude screening for TDCP: restricting TDCP processing to higher-altitude rays improves consistency with the Gaussian noise assumptions of the filter by making residual dispersive delays small relative to thermal noise.

\begin{table}[ht!]
    \centering
    \caption{TDCP measurement error statistics for different ranges of tangential altitudes. Samples with absolute error greater than 10 m are excluded from the mean and standard deviation computation.}
    \begin{tabular}{c|c c|c c| c c}
        \hline \hline
        \multirow{2}{*}{Tangential Altitude} & \multicolumn{2}{c|}{Receiver Noise [mm]} & \multicolumn{2}{c|}{Ephemeris Error [mm]} & \multicolumn{2}{c}{Ionospheric Delay [mm]} \\
        \cline{2-7}
         & Mean & Std. Dev. & Mean & Std. Dev. & Mean & Std. Dev. \\    
        \hline
        0--1000 km      & 0.457 & 0.108 &  0.042 & 1.619 & 0.100 & 3.451 \\
        1000--5000 km  & 1.634 & 1.181 & -0.025 & 1.519 & 0.015 & 3.093 \\
        5000--20000 km & 3.800 & 1.246 &  0.010 & 1.374 & 0.011 & 0.342 \\
        \hline \hline
    \end{tabular}
    \label{tab:tdcp_error_stats}
\end{table}

\subsection{Summary of Section}
\label{sec:measurement_summary}
This section quantified the principal error sources affecting terrestrial GNSS navigation at lunar distances. The results expose a fundamental trade-off governed by tangential altitude: higher-altitude rays reduce ionospheric/plasmaspheric residuals but also reduce $C/N_0$, increasing thermal noise.

Pseudorange measurements offer the highest availability but are strongly affected by ionospheric and plasmaspheric delays, particularly at low-to-medium tangential altitudes. Using single-frequency pseudorange without introducing significant biases requires one (or more) of the following: (i) aggressive tangential-altitude screening, (ii) inflating measurement variances (suboptimal but simple), (iii) compensation using external delay models, or (iv) explicit estimation (or addition as consider states where cross-covariance is taken into account but the states are not explicitly estimated~\cite{Tapley2004}) of delay states in the filter. 
Ionosphere-free pseudorange mitigates the dominant first-order delay but reduces availability and increases receiver noise. 
TDCP provides a favorable compromise when tangential-altitude screening is applied, because differencing removes integer ambiguity and suppresses slowly varying biases while retaining mm-level precision when the ionospheric delays are small compared to receiver noise.

Table~\ref{tab:measurement_tradeoff} summarizes the qualitative trade-offs observed in the simulation and the corresponding preferable tangential-altitude screening strategy for maintaining delay residuals smaller than the receiver noise.

\begin{table}[ht!]
    \centering
    \caption{Trade-off analysis of measurement types and preferable tangential altitude thresholding to keep biases from ionospheric and plasmaspheric delays small compared to receiver noise for lunar GNSS navigation.}
    \begin{tabular}{c|c|c|c|c}
        \hline \hline
        Measurement Type & Availability & Receiver Noise & Iono Delay & Preferable Altitude Threshold \\
        \hline
        Pseudorange (L1) & High & Medium & Large & High ($>$10000 km) \\
        Pseudorange (L5) & High & Low & Very Large & Very High ($>$15000 km) \\
        Iono-free Pseudorange & Medium & High & Medium & Low ($>$1000 km) \\
        TDCP (L1) & Medium & Medium & Medium & Medium ($>$5000 km)\\
        \hline \hline
    \end{tabular}
    \label{tab:measurement_tradeoff}
\end{table}

%% file: sections/7_result.tex
\section{Simulation Results}
\label{sec:result}

\subsection{Filter Settings}
\label{sec:filter_settings}
We evaluate the proposed UDU filter and fixed-interval smoother under three measurement configurations:
(1) L1 pseudorange only, (2) ionosphere-free pseudorange only, and (3) ionosphere-free pseudorange combined with TDCP.
Configuration (1) serves as a baseline and follows common assumptions in prior lunar GNSS ODTS studies~\cite{Iiyama2024TDCPNavigation, Mina2025LCRNS}: we process L1 pseudorange with a tangential-altitude mask of 1000~km to reject rays that traverse the ionosphere.
Configuration (2) replaces single-frequency pseudorange with ionosphere-free pseudorange using the same 1000~km mask to further suppress higher-order ionospheric terms and bending effects that are not removed by the first-order ionosphere-free combination.
Configuration (3) combines ionosphere-free pseudorange (1000~km mask) with TDCP measurements, applying a stricter mask of 5000~km for TDCP to preferentially retain measurements for which time-differenced ionospheric/plasmaspheric residuals are small relative to carrier-phase noise.

To test robustness against cycle slips, we inject random cycle slips in the TDCP: for measurements with $(C/N_0)<25$~dB-Hz, each TDCP measurement experiences a slip with probability 10\%.

Based on the measurement error analysis in Section~\ref{sec:meas_error_analysis}, we set the broadcast-ephemeris-induced range error standard deviation to $\sigma_{\text{URE}}=\SI{10.0}{m}$ for pseudorange and the time-differenced equivalent to $\sigma_{\Delta \text{URE}}=\SI{8}{mm}$ for TDCP. The altitude masks and these inflation terms were selected from multiple trial runs; further tuning may yield additional improvement.

The initial 1$\sigma$ uncertainties are set to \SI{1000}{m} for position and \SI{1}{m/s} for velocity. For the clock, the initial standard deviations are \SI{1000}{m} for clock bias, \SI{1}{m/s} for clock drift, and \SI{1}{mm/s^2} for clock acceleration. The initial standard deviation of the SRP coefficient is set to 20\% of its true value. The initial covariance is diagonal with entries equal to the squared initial standard deviations. The unmodeled acceleration process noise is modeled as isotropic white acceleration with spectral density $q_a=(\SI{1e-9}{m/s^2})^2$ per axis.

For each measurement configuration, we run 40 Monte-Carlo simulations with independent realizations of the initial state perturbations and measurement errors. To ensure convergence from the large initial uncertainties, performance is evaluated over the final orbit revolution (approximately 30 hours). We report SISE metrics consistent with the LunaNet SRD~\cite{nasa2022srd} as follows:
\begin{align}
    SISE_{\text{pos}} &=
    \sqrt{ (\hat{r}_x-r_x)^2 + (\hat{r}_y-r_y)^2 + (\hat{r}_z-r_z)^2 + c^2(\hat{\delta t}-\delta t)^2 }, \\
    SISE_{\text{vel}} &=
    \sqrt{ (\hat{v}_x-v_x)^2 + (\hat{v}_y-v_y)^2 + (\hat{v}_z-v_z)^2 + c^2(\hat{\delta \dot{t}}-\delta \dot{t})^2 } .
\end{align}

\subsection{Results}
\label{sec:simulation_results}
Table~\ref{tab:filtering_results} summarizes the position and velocity SISE statistics (RMS, 95\%, and 99.7\% percentiles) for the filter and smoother, averaged over 40 Monte-Carlo runs.

\begin{table}[ht!]
    \centering
    \caption{SISE position and velocity errors of the filter and smoother at the final orbit under different measurement configurations. Values represent are computed over 40 Monte Carlo simulations (PR: pseudorange, TDCP: time-differenced carrier phase, IF: ionosphere-free). Using L1 pseudorange only results in the largest SISE errors due to unmodeled plasmaspheric delays. The combination of ionosphere-free pseudorange and TDCP achieves the smallest SISE errors. Application of smoothing reduces the 95th- and 99.7th-percentile SISE errors.}
    \begin{tabular}{c|c|c| c |c|c|c|c|c|c}
        \hline \hline
        \multicolumn{3}{c|}{Measurements (Altitude Mask)} & \multirow{2}{*}{Method} &
        \multicolumn{3}{c|}{Position SISE (m)} & \multicolumn{3}{c}{Velocity SISE (mm/s)} \\
        \cline{1-3} \cline{5-10}
        L1 PR & IF PR & L1 TDCP & & RMS & 95\% & 99.7\% & RMS & 95\% & 99.7\% \\
        \hline
        \checkmark      &  &  & Filter   & 18.41 & 28.73 & 31.31 & 0.82 & 1.48 & 2.11 \\
       ($\geq$ 1000 km) &  &  & Smoother & 10.84 & 14.26 & 14.85 & 0.70 & 1.58 & 2.13 \\
        \hline
         & \checkmark       &  & Filter   &  6.05 &  9.74 & 12.39 & 0.46 & 0.91 & 1.80 \\
         & ($\geq$ 1000 km) &  & Smoother &  4.77 &  7.71 &  9.65 & 0.36 & 0.74 & 1.54 \\
        \hline
         & \checkmark & \checkmark & Filter   &  4.53 &  7.44 & 10.20 & 0.45 & 0.98 & 2.08 \\
         & ($\geq$ 1000 km) & ($\geq$ 5000 km) & Smoother &  2.79 &  4.47 &  7.31 & 0.30 & 0.64 & 1.46 \\
        \hline \hline
    \end{tabular}
    \label{tab:filtering_results}
\end{table}

\subsubsection{L1 Pseudorange Only (Baseline)}
Using only L1 pseudorange yields the largest SISE errors for both position and velocity. This is consistent with unmodeled plasmaspheric delays contaminating pseudorange even after applying the 1000~km tangential-altitude mask. These results suggest that pseudorange-only filtering requires either (i) more aggressive variance inflation to maintain consistency or, more optimally, (ii) explicit modeling/estimation of additional bias states (e.g., plasmaspheric delay and LOS range biases) to prevent systematic errors. 
For single-frequency receivers, an alternative is the Group and Phase Ionosphere Correction (GRAPHIC) observable~\cite{Bock2009GPS_SF_Orbit, Shi2012IonosphericPPP}, which suppresses first-order plasmaspheric delay by combining code and phase; however, this requires estimating float ambiguities, increasing state dimension and implementation complexity.

\subsubsection{Ionosphere-free Pseudorange Only}
Replacing single-frequency pseudorange with ionosphere-free pseudorange substantially improves both position and velocity SISE, despite reduced availability (dual-frequency tracking is required, and fewer satellites transmit both signals). This improvement indicates that first-order plasmaspheric delays are a dominant error source for the baseline configuration, and that ionosphere-free measurements provide a practical mitigation in the absence of an explicit plasmaspheric delay estimator.

\subsubsection{Ionosphere-free Pseudorange + TDCP}
Adding TDCP to the ionosphere-free pseudorange improves the overall SISE performance for both the filter and smoother. For the filter, the position RMS decreases from 6.05m (IF PR only) to 4.53m (IF PR + TDCP), corresponding to an improvement of approximately 25\%. The position 95\% and 99.7\% SISE are also reduced, from 9.74m and 12.39m to 7.44m and 10.20m, respectively. These results confirm that TDCP provides strong relative position and clock bias constraints between successive epochs, particularly benefiting the position solution.

\begin{figure}[ht!]
    \centering
    \includegraphics[width=\textwidth]{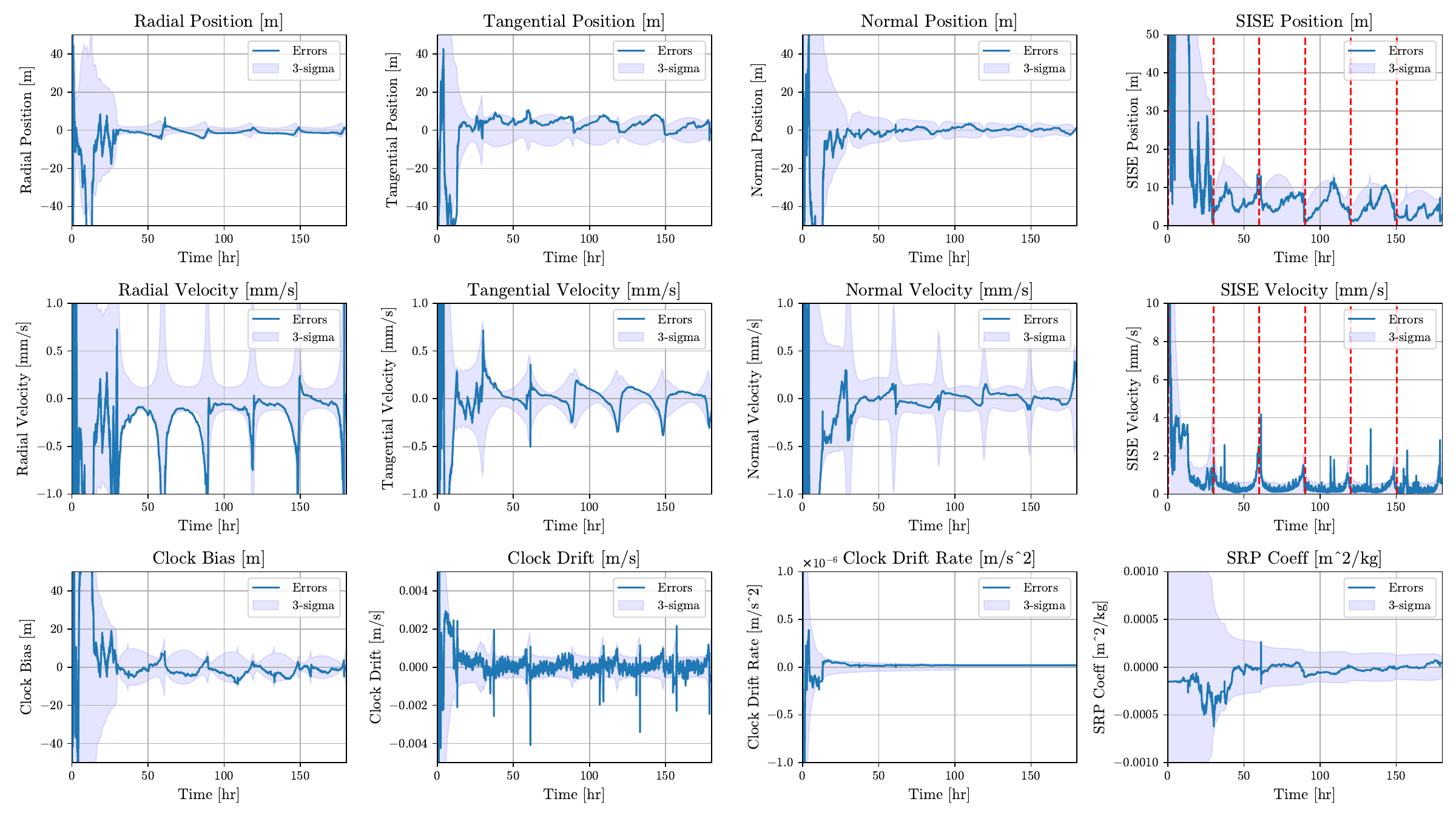}
    \caption{State estimation errors and 3$\sigma$ bounds, and the corresponding SISE position and velocity errors, for one Monte-Carlo run using iono-free pseudorange and TDCP. The red vertical line indicates the periapsis passes.}
    \label{fig:filtering_error}
\end{figure}

For velocity, the filter RMS remains nearly unchanged (0.46mm/s vs.\ 0.45mm/s), but the high-percentile performance degrades. The 95\% velocity SISE increases from 0.91mm/s to 0.98mm/s, and the 99.7\% value increases from 1.80mm/s to 2.08mm/s. This behavior is consistent with occasional near-outlier TDCP measurements that transiently perturb the velocity and clock-drift states. 
Since the TDCP measurements have low measurement errors, one near-outlier measurement can largely perturb the velocity estimates.
As shown in Fig.~\ref{fig:filtering_error}, these events appear as isolated spikes in the velocity and clock-drift errors. Outside such events, the estimation errors largely remain within the 3$\sigma$ bounds, indicating overall filter consistency under the adopted inflation and screening strategy.

\subsubsection{Effect of Smoothing}
Applying the fixed-interval smoother improves both position and velocity SISE across all measurement configurations. The largest benefit is observed when TDCP is included. With ionosphere-free pseudorange only, smoothing reduces the position RMS from 6.05m to 4.77m and the velocity RMS from 0.46mm/s to 0.36mm/s. When TDCP is added, the improvements are more pronounced: the position RMS decreases from 4.53m to 2.79m (approximately 38\% reduction), and the velocity RMS decreases from 0.45mm/s to 0.30mm/s (approximately 33\% reduction).

\begin{figure}[ht!]
    \centering
    \includegraphics[width=0.9\textwidth]{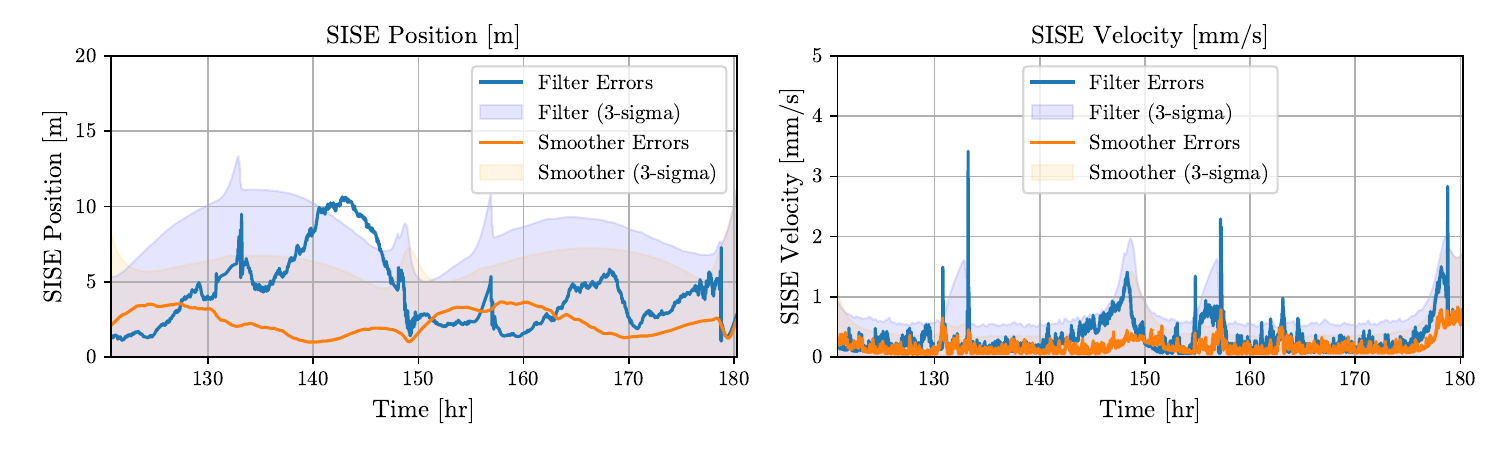}
    \caption{Filter (blue) and smoother (orange) errors and 3$\sigma$ bounds for position, velocity, clock bias, and clock drift for one Monte-Carlo run using ionosphere-free pseudorange and TDCP. The smoother result is shown after five filter--smoother iterations.}
    \label{fig:smoothing_error}
\end{figure}

Figure~\ref{fig:smoothing_error} compares the filter and smoother trajectories (after five filter--smoother iterations) for one Monte-Carlo run with ionosphere-free pseudorange and TDCP. Large position SISE errors near apolune, where slower orbital motion reduces instantaneous observability, and larger velocity SISE errors near perilune are both noticeably reduced after smoothing.

Smoothing also mitigates the high-percentile degradation introduced by TDCP in the filter. In particular, the 99.7\% velocity SISE decreases from 2.08mm/s (filter) to 1.46mm/s (smoother), a reduction of about 30\%. Compared to the ionosphere-free pseudorange-only smoother (1.54mm/s at 99.7\%), the TDCP-assisted smoother achieves slightly lower extreme velocity errors, while providing substantially improved position accuracy (2.79m vs.\ 4.77~m RMS). These results indicate that the backward information flow in the smoother effectively suppresses transient TDCP-induced spikes and yields the best overall consistency and accuracy among the tested configurations.

\subsubsection{Estimation Error Distribution}
\begin{figure}[ht!]
    \centering
    \includegraphics[width=0.8\textwidth]{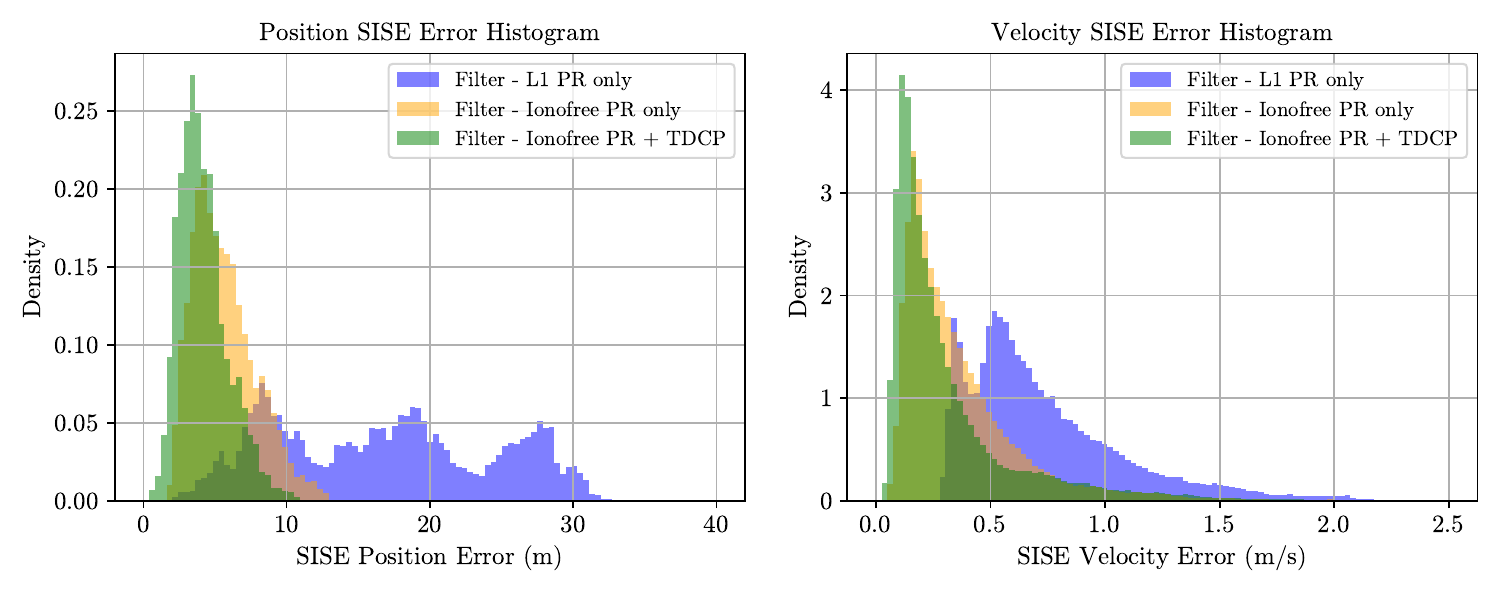}
    \caption{Histogram of filter position and velocity SISE over 50 Monte-Carlo simulations (final orbit only).}
    \label{fig:error_histogram}
\end{figure}

\begin{figure}[ht!]
    \centering
    \includegraphics[width=0.8\textwidth]{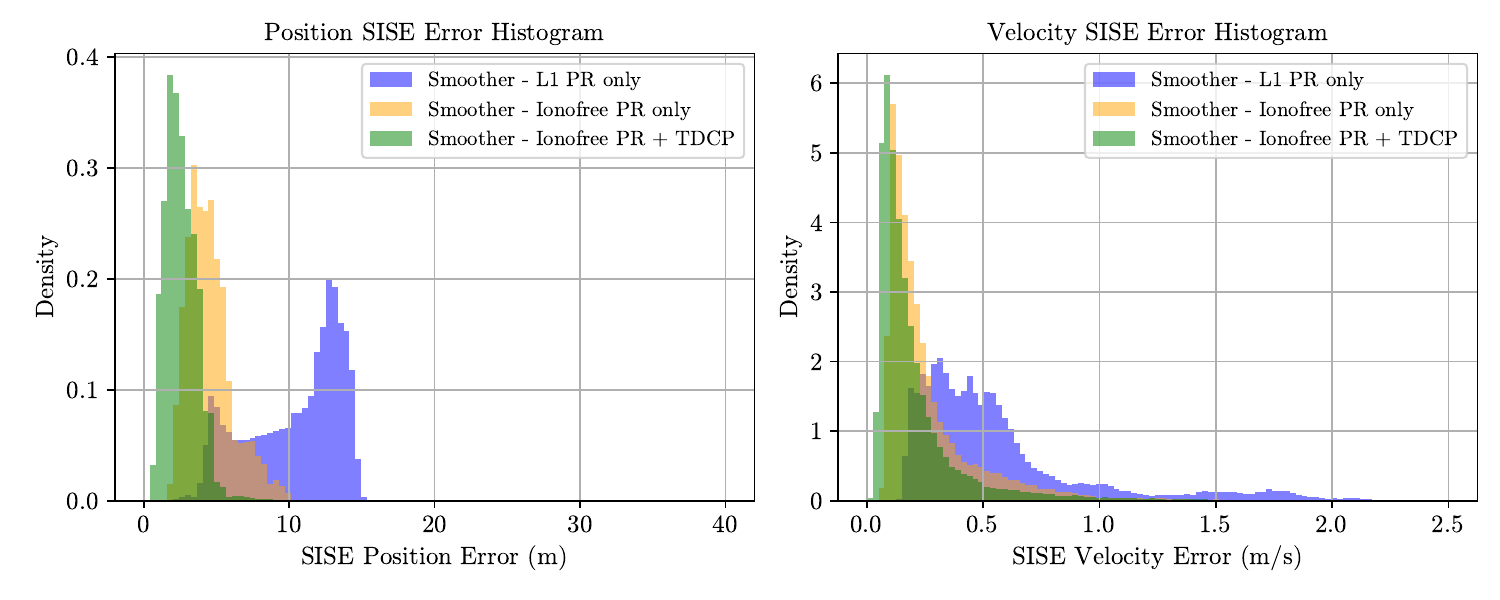}
    \caption{Histogram of smoother position and velocity SISE over 50 Monte-Carlo simulations (final orbit only).}
    \label{fig:smoother_error_histogram}
\end{figure}

These trends are also reflected in the SISE histograms over the final orbit across 40 Monte-Carlo runs (Figs.~\ref{fig:error_histogram} and~\ref{fig:smoother_error_histogram}). Incorporating TDCP shifts both the position and velocity SISE distributions toward smaller values for both filtering and smoothing. Comparison of filter versus smoother results shows that smoothing reduces the tails of the SISE distributions, consistent with its ability to incorporate future measurements to refine poorly observed portions of the orbit.

%% file: sections/8_conclusion.tex
\section{Conclusion}
\label{sec:conclusion}
This paper presented a terrestrial GNSS–based orbit and clock estimation framework for lunar navigation satellites that combines stochastic cloning with a numerically robust UD-factorized filter and a delayed-state fixed-interval smoother. 
To address the low observability and numerical sensitivity that arise when incorporating precise TDCP measurements at lunar distances, we developed UD-factorization filter that can process delayed-state measurements with stochastic cloning. 
In addition, we formulated dynamics and measurement models that explicitly include relativistic coupling between orbit and clock states, transformations among Earth and lunar time scales, and propagation delays through the terrestrial ionosphere and plasmasphere.

We evaluated the proposed approach using high-fidelity Monte-Carlo simulations that incorporate realistic multi-constellation GNSS geometry, broadcast-ephemeris errors, and ray-tracing-based ionospheric and plasmaspheric delay models. The results show that augmenting ionosphere-free pseudorange with TDCP yields a clear improvement in estimation performance relative to pseudorange-only baselines. For filtering, the inclusion of TDCP reduces the RMS position SISE from \SI{6.05}{m} to \SI{4.53}{m}, corresponding to an improvement of approximately 25\%, while the RMS velocity SISE remains comparable at about \SI{0.45}{mm/s}.

The delayed-state smoother provides additional gains by reducing both the mean error levels and the tails of the error distributions. With ionosphere-free pseudorange and TDCP, the smoother achieves an RMS position SISE of \SI{2.79}{m} and an RMS velocity SISE of \SI{0.30}{mm/s}, while substantially suppressing high-percentile velocity errors induced by occasional TDCP outliers. These results demonstrate that the combined use of TDCP and smoothing enables sub-\SI{3}{m} position accuracy and sub-\si{mm/s} velocity accuracy, meeting the position accuracy target envisioned for LANS.

Future work will focus on three main directions. First, we will pursue systematic tuning of tangential-altitude masking thresholds and measurement noise inflation to more effectively balance measurement availability against biases induced by plasmaspheric delays. Second, we will reduce reliance on masking by explicitly estimating residual ionospheric and plasmaspheric delay terms, as well as multi-constellation inter-system biases, either directly within the state vector or as consider states. Third, we will target further improvements in velocity and clock-drift estimation to meet the LCRNS 3$\sigma$ SISE velocity requirements (1.2 mm/s), through the incorporation of additional measurements such as inter-satellite ranging and occasional ground-station tracking, along with continued filter and smoother refinement. In addition, we will investigate the direct processing of raw pseudorange and GRAPHIC observables with explicit ambiguity handling, with the goal of improving robustness and availability in single-frequency or intermittently dual-frequency reception scenarios.